%% file: main.tex
\documentclass[10pt]{article} % For LaTeX2e
\usepackage[a4paper, total={6.5in, 10in}]{geometry}

\input{math_commands.tex}

\usepackage{hyperref}
\usepackage{subcaption}
\usepackage{graphicx} 
\usepackage{booktabs}
\usepackage{tabularx}
\usepackage{url}
\usepackage{xcolor}

\title{ASTRA: A Scene-aware TRAnsformer-based model for trajectory prediction}

\author{Izzeddin Teeti\thanks{Visual Artificial Intelligence Laboratory, Oxford Brookes University, UK} 
\and Aniket Thomas\thanks{Indian Institute of Technology Bombay, India} 
\and Munish Monga\footnotemark[2]
\and Sachin Kumar\footnotemark[2]
\and Uddeshya Singh\footnotemark[2]
\and Andrew Bradley\footnotemark[1]
\and Biplab Banerjee\footnotemark[2] 
\and Fabio Cuzzolin\footnotemark[1] 
\and \tt\small iteeti@brookes.ac.uk,
}

\begin{document}

\maketitle

\begin{abstract}

% update the figure 1, Aniket \\
% Add non Unet metrics in table 2, Aniket \\
% need to change ASAP to ASTRA in the paper, Izzeddin (Done)\\
% write more about the graph, Izzeddin (Done)\\
% add emphasis on Scene dimension, in Conclusslion write instead of U-Net we could have used Transformer based AutoEncoder but scene dimension is important, Aniket \\
% more attention on light weight, Uddeshya(done) \\

We present ASTRA (\textbf{A} \textbf{S}cene-aware \textbf{TRA}nsformer-based model for trajectory prediction), a light-weight pedestrian trajectory forecasting model that integrates the scene context, spatial dynamics, social inter-agent interactions and temporal progressions for precise forecasting. We utilised a U-Net-based feature extractor, via its latent vector representation, to capture scene representations and a graph-aware transformer encoder for capturing social interactions. These components are integrated to learn an agent-scene aware embedding, enabling the model to learn spatial dynamics and forecast the future trajectory of pedestrians.
% ASTRA utilizes a U-Net based feature extractor, graph-aware and scene-aware transformer encoders that contribute to generating an agent-scene aware embedding. 
The model is designed to produce both deterministic and stochastic outcomes, with the stochastic predictions being generated by incorporating a Conditional Variational Auto-Encoder (CVAE).
% A weighted penalty loss function is utilized by ASAP
ASTRA also proposes a simple yet effective weighted penalty loss function, which helps to yield predictions that outperform a wide array of state-of-the-art deterministic and generative models. ASTRA demonstrates an average improvement of 27\%/10\% in deterministic/stochastic settings on the ETH-UCY dataset, and 26\% improvement on the PIE dataset, respectively, along with seven times fewer parameters than the existing state-of-the-art model (see Figure \ref{fig:param-compare}). Additionally, the model's versatility allows it to generalize across different perspectives, such as Bird's Eye View (BEV) and Ego-Vehicle View (EVV).
\end{abstract}

\begin{figure}[h]
\vspace{-0.1cm}
\begin{center}
\includegraphics[width=0.8\linewidth, height=6cm]{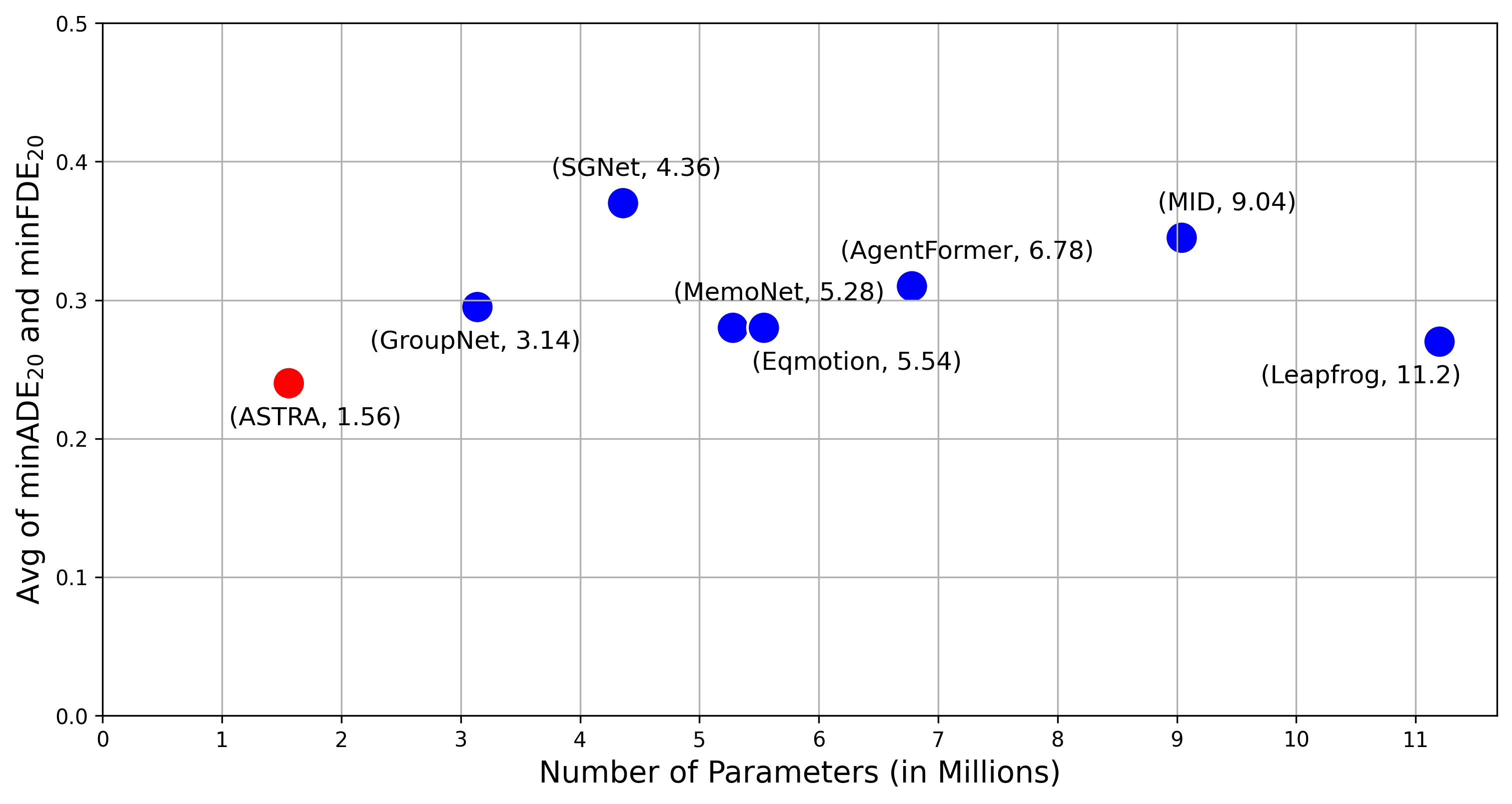}
\end{center}
\caption{Comparison of average (minADE\textsubscript{20}/minFDE\textsubscript{20}) against the number of parameters for various models on the ETH-UCY dataset. Each point represents a different model, with the model name and number of parameters in millions indicated. Our model, ASTRA, achieves state-of-the-art results with the least number of parameters, demonstrating its efficiency and effectiveness in pedestrian trajectory forecasting.}
\label{fig:param-compare}
\end{figure}

% \begin{figure}
%     \vspace{-1.10cm}
%     \centering
%     \includegraphics[width=\linewidth, height=5.5cm]{Figures/model_sizes_comparison.png} % Adjust width and height as needed
%     \captionof{figure}{Comparison of average (minADE\textsubscript{20}/minFDE\textsubscript{20}) against the number of parameters for various models on the ETH-UCY dataset. Each point represents a different model, with the model name and number of parameters in millions indicated. Our model, ASAP, achieves state-of-the-art results with the least number of parameters, demonstrating its efficiency and effectiveness in pedestrian trajectory forecasting.}
%     \label{fig:param-compare}
% \end{figure}

\section{Introduction}
\label{sec:intro}

\begin{figure}[ht!]
    \centering
    \begin{subfigure}{0.4\textwidth}
        \centering
        \includegraphics[width=\textwidth]{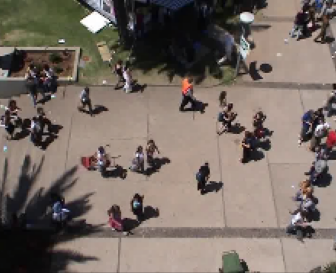}
        \caption{ETH dataset sample}
        \label{fig:eth_dataset_sample}
    \end{subfigure}
    \hspace{1mm}
    \begin{subfigure}{0.4\textwidth}
        \centering
        \includegraphics[width=\textwidth]{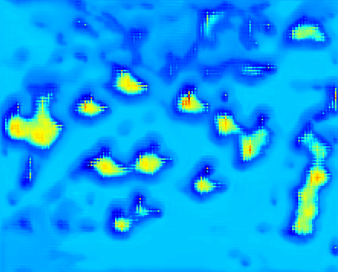}
        \caption{Grad-CAM of ETH sample}
        \label{fig:eth_gradcam}
    \end{subfigure}\\
    \begin{subfigure}{0.4\textwidth}
        \centering
        \includegraphics[width=\textwidth]{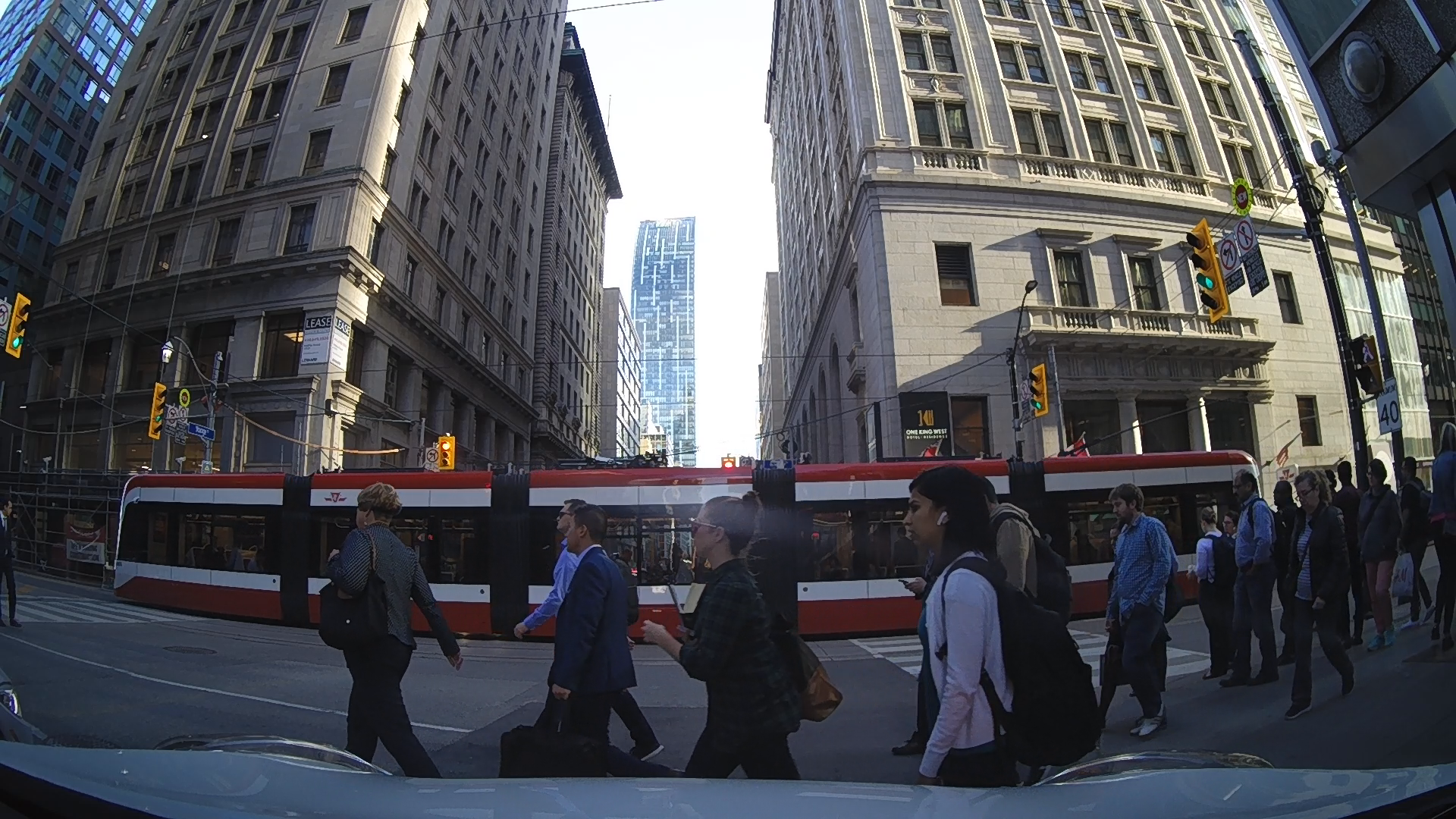}
        \caption{PIE dataset sample}
        \label{fig:pie_dataset_sample}
    \end{subfigure}
    \hspace{1mm}
    \begin{subfigure}{0.4\textwidth}
        \centering
        \includegraphics[width=\textwidth]{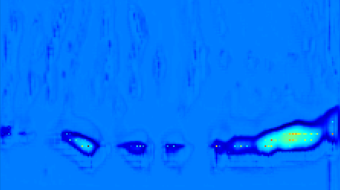}
        \caption{Grad-CAM of PIE sample}
        \label{fig:pie_gradcam}
    \end{subfigure}
    \caption{Sample images from the BEV dataset (ETH), and EVV dataset (PIE), along with their Grad-CAM representation from U-Net.}
    \label{fig:datasets}
\end{figure}

The pursuit of forecasting human trajectories is central, acting as a cornerstone for devising secure and interactive autonomous systems across various sectors. This endeavour is crucial in a wide array of applications, encompassing autonomous vehicles, drones, surveillance, human-robot interaction, and social robotics.
Furthermore, it is crucial for predictive models to strike a balance between accuracy, dependability, and computational efficiency, given the imperative for these models to function on in-vehicle processing units with limited capabilities. The challenge of trajectory prediction involves estimating the future locations of agents within a scene, given its past trajectory. This estimation task can be tackled either through Bird's Eye View (BEV) (Figure \ref{fig:eth_dataset_sample}) or Ego-Vehicle View (EVV) (Figure \ref{fig:pie_dataset_sample}) perspectives. This demands a comprehensive understanding of the scene, in addition to spatial, temporal, and social aspects that govern human movement and interaction.

To solve the prediction problem, various building blocks, including RNNs, 3D-CNNs, and transformers, have been employed to address the temporal dimension, with transformers demonstrating superior efficacy \cite{giuliari2021transformer, rasouli2023pedformer}. However, temporal modelling alone is unaware of the social behaviour of the agents within the scene, i.e. how agents interact with one another. In addressing the social dimension, methods such as Social Pooling \cite{Pang2021} and Graph Neural Networks \cite{GNN} (GNNs) have been explored, with GNN emerging as the most effective \cite{EqMotion}. Some researchers have integrated both transformers and GNN, either sequentially or in parallel, to refine the prediction paradigm \cite{li2022graph, Chen2023vnagt, jia2023hdgt, liu2022social}. However, these approaches entail heightened computational burdens due to the resource-intensive nature of both GNNs and transformers. Furthermore, transformers, by their inherent design, may pose potential challenges in preserving information, as they do not inherently accommodate the graph structure in their input. On the other hand, scene dimension, or scene embedding, delves into the interaction between an agent and its surroundings. \cite{Rasouli2021} and \cite{Mangalam2021} utilised semantic segmentation maps which enhanced the model’s grasp of environmental context. Another aspect across all surveyed papers, however, is their tendency to focus exclusively on either BEV or EVV, rarely considering both like \cite{yao2021bitrap}. This narrow focus becomes particularly problematic in, e.g., unstructured environments where a BEV might not be available, limiting the applicability of these methods.

In light of these challenges, this paper introduces a lightweight model, coined ASTRA (\textbf{A} \textbf{S}cene-aware \textbf{TRA}nsformer-based model for trajectory prediction). By integrating a U-Net-based key-point extractor \cite{ribera2019locating}, ASTRA captures essential scene features without relying on explicit segmentation map annotations and alleviates the data requirements and preprocessing efforts highlighted earlier. This method also synergises the strengths of GNNs in representing the social dimension of the problem and of transformers in encoding its temporal dimension. Crucially, our approach processes spatial, temporal, and social dimensions concurrently, by embedding the graph structure into the token's sequence prior to the attention mechanism, rendering the transformer graph-aware. 
The model does so while keeping the complexity of the model minimal. To refine the model's ability to accurately learn trajectories, we implemented a modified version of the trajectory prediction loss, incorporating a penalty component. This is in contrast to \cite{AgentFormer} which does not build a graph and does not preserve the social structure; they distinguish between self-agent and all other agents, then they treat all other agents the same without encoding the positional or structural encodings.

% \Munish{\textbf{Agentformer Comparison}: Agentformer does not build a graph and does not preserve the social structure; they distinguish between self-agent and all other agents, they treat all other agents the same without encoding the positional or structural encodings, also they encode the scene using semantic maps. In contrast, the scene encodings given by the U-Net based keypoint extractor in our model capture a global scene context, that additionally provides social interaction among agents in a frame.}

Furthermore, distinct from the vast majority of models in this domain, our model demonstrates generalisability by being applicable to both types of trajectory prediction datasets, BEV and EVV.

Our methodology underwent evaluation using renowned benchmark trajectory prediction datasets (Figure \ref{fig:datasets}): ETH \cite{eth2009}, UCY \cite{ucy2007}, and the PIE dataset \cite{piedataset}. The empirical findings highlight ASTRA's outperforming the latest state-of-the-art methodologies. Notably, our method showcased significant improvements of 27\% on the deterministic and 10\% on the stochastic settings of the ETH and UCY datasets and 26\% on PIE. 

While maintaining high accuracy, ASTRA also features a significant reduction in the number of model's parameters (Figure \ref{fig:param-compare}) - seven times fewer than the existing competing state-of-the-art model \cite{leapfrog}. 

The paper's highlights are as follows:
\begin{enumerate}

    \item A lightweight model architecture that is seven times lighter than the existing state-of-the-art model, tailored for deployment on devices with limited processing capabilities while maintaining state-of-the-art predictive performance.
        
    \item A loss-penalization strategy that enhances trajectory prediction, featuring a weighting trajectory loss function that dynamically adjusts penalty progression in response to prediction challenges.

    \item Utilisation of the Scene-aware embeddings with a U-Net-based feature extractor to encode scene representations from frames, addressing a critical aspect often overlooked in recent works.
    
    \item A graph-aware transformer encoder that contributes significantly to generating Agent-Scene aware embedding for improved prediction accuracy, ensuring informed inter-agent interaction capture.
    
    % \item Utilisation of the U-Net based feature extractor encoder and a graph-aware transformer encoder that contributes significantly to generating Agent-Scene aware embedding for improved prediction accuracy, ensuring informed inter-agent interaction capture.

\end{enumerate}

\section{Related Work}
\subsection{Trajectory Prediction}
The trajectory prediction problem is usually approached in two ways: stochastic (multi-modal) predictions  \cite{leapfrog, AgentFormer, memonet, yao2021bitrap} and deterministic (uni-modal) predictions \cite{socialforce,pellegrini2009you,S-LSTM}. The model produces only one prediction (most probable) per input motion in a deterministic setting. In contrast, a stochastic setting involves the model generating multiple predictions for each input motion. Stochastic approaches utilize generative techniques like Conditional Variational Auto-Encoders (CVAEs) \cite{AgentFormer, yao2021bitrap}, Generative Adversarial Networks (GANs) \cite{huang2021sti}, Normalizing Flows \cite{nflow}, or Denoising Diffusion Probabilistic Models \cite{leapfrog} to introduce randomness into the prediction process, thereby generating diverse future trajectories with varying qualities for each pedestrian, aiming to elucidate the distribution of potential future coordinates of pedestrian trajectories. ASTRA also offers both deterministic and stochastic predictions like some of the previous works \cite{EqMotion, yao2021bitrap, Traj++}.

\subsection{Social and Scene Dimension}
The social dimension focuses on capturing agent-agent interactions, emphasizing how individuals or objects influence each other's movements within a shared space. Notably, some methodologies incorporate social pooling, concurrently with attention mechanisms \cite{Pang2021}. Algorithms in this domain predominantly leverage various forms of Graph Neural Networks (GNNs) to encapsulate the social dynamics among agents. Some methods employ a fully connected undirected graph, encompassing all scene agents \cite{EqMotion, gohome2021, socialbigat}. This approach, albeit comprehensive, escalates exponentially with the number of agents (nodes). Conversely, other methods opt for sparsely connected graphs, establishing connections solely among agents within a proximal range, thereby reducing the linkage count substantially \cite{dada2021, girase2021loki, Traj++, Weng2021ptp}. In a similar vein, \cite{AgentFormer, Traj++} proposes sparse, directional graphs, predicated on the premise that different agent types possess varying perceptual ranges. Regarding the optimal depth of GNN layers, \cite{Addanki2021} advocate for deeper graphs to enhance performance. This stands in contrast to the findings of \cite{Weng2021ptp} and \cite{liu2020spatiotemporal}, who posit that two layers are optimal. Nevertheless, this depth increases computational demands, particularly when agent nodes are numerous, posing challenges for autonomous vehicle applications reliant on edge devices for processing.

The scene dimension, extracted from the video frames, includes the low-level representation of the physical environment, obstacles, and any elements that could affect the agent's path, ensuring a comprehensive understanding of both social and environmental factors in predicting movement trajectories. 
\cite{Rasouli2021} and \cite{Mangalam2021} capture scene dimension with the help of semantic segmentation to delineate visual attributes of varied classes, subsequently elucidating their interrelations via attention. However, obtaining a panoptic segmentation mask, might not be always feasible. Also, this approach introduces a considerable dependency on the availability of additional segmentation maps, presenting a challenge in terms of data requirements and preprocessing efforts.

\subsection{Temporal Dimension}
Understanding the trajectory history of an agent significantly augments the predictive accuracy regarding its potential future path. Predominantly, ego-camera-based models are tailored to shorter temporal horizons and employ 3D Convolutional Neural Networks (3D-CNNs) \cite{dada2021, Kotseruba2021}. Some research, instead, adopts Hidden Markov Models (HMMs) for temporal analysis \cite{Cai2020}. For scenarios necessitating extended time horizon considerations, more intricate structures are proposed, including Transformers \cite{AgentFormer, EqMotion, transf, chen2021s2tnet} and various forms of Recurrent Neural Networks (RNNs) \cite{girase2021loki}, including
Long Short-Term Memory networks (LSTMs) 
\cite{rasouli2023pedformer, Bhattacharyya2021, dada2021} and Gated Recurrent Units (GRUs) \cite{gohome2021}. Both Transformer and RNN-based models have exhibited superior performance, often achieving state-of-the-art results in this domain. However, some of these models tend to address the temporal dimension in isolation from the social context. This segregated approach potentially results in information loss and contributes to an increased computational load, necessitated by the addition of separate components to handle the social dimension. Consequently, there emerges a pressing demand for integrated models capable of concurrently processing both temporal and social dimensions. A promising direction in this regard is the development of graph-aware transformers, which encapsulate the essence of both temporal dynamics and social interactions within a unified framework. 

\subsection{Graph-aware Transformers}
Graph-aware transformers aim to compound the benefits of graphs (with their associated social embeddings) and of transformers, with their acclaimed attention mechanism and temporal embeddings. Notably, these advancements have predominantly catered to graph-centric datasets like ACTOR \cite{Tang2009} and CHAMELEON SQUIRREL \cite{rozemberczki2021multi}. Direct application of graph-aware transformers remains untouched in pedestrian trajectory forecasting, with prevalent methodologies leaning towards transformers processing embeddings emanating from graphs \cite{li2022graph, Chen2023vnagt, jia2023hdgt}. There has been a discernible preference for using GNN and transformer blocks, rather than fully-integrated graph-aware transformers.

A comprehensive evaluation of numerous contemporary graph-transformer models across three graph-centric datasets is conducted in \cite{muller2023attending}. The analysis reveals a consistent pattern: models employing Random Walk for structural encoding exhibit superior performance across all tested datasets. Building on this empirical evidence, our approach utilizes Random Walk to encode the pedestrian graph, which is then seamlessly integrated into the transformer architecture. This integration is designed to yield a graph-aware transformer, thereby enhancing the model's capability to effectively capture and interpret complex pedestrian dynamics within various environments. To the best of the authors' knowledge, this is the first work towards utilising a graph-aware transformer to solve the trajectory prediction problem, opposing many methods which use graphs along with transformers.

\section{Methodology}
\label{sec:method}

\begin{figure*}[ht!]
    \vspace{-0.1cm}
    \centering
    \includegraphics[width=\linewidth]{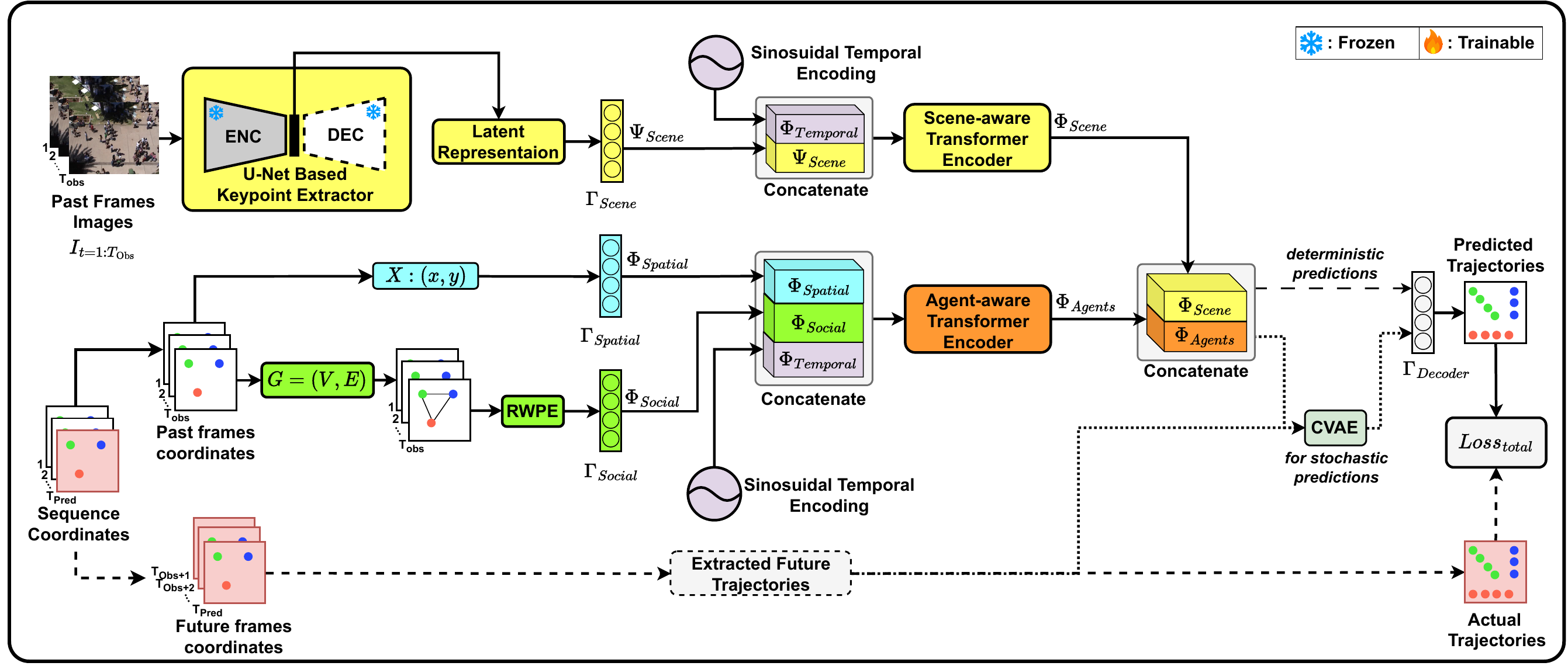}
    \caption{\small \textbf{Model Architecture.}
    Overview of ASTRA model architecture for pedestrian trajectory forecasting.}
    \label{fig:model}
\end{figure*}
\textbf{Problem Formulation.} The objective of pedestrian trajectory prediction is to forecast the future position of a pedestrian based on the observed historical sequence of the pedestrian's positions. For this, the historical sequence of the pedestrians is provided as a sequence of coordinates $\boldsymbol{X} = \{X^{a}_{t} \ | \ t \in \left(1,2, \ldots ,T_{obs}\right)$ $; a \in \left(1,2,\ldots,A\right)\}$ related to $A$ target agents extracted over the previous $T_{obs}$ time instants.
Here $X^{a}_{t}$ is a pair of 2D coordinates $\{x^{a}_{t}, y^{a}_{t}\}$ for BEV datasets (see Figure \ref{fig:eth_dataset_sample}), and a set of bounding box coordinates, $\{x^{a}_{1,t}, y^{a}_{1,t}, x^{a}_{2,t}, y^{a}_{2,t}\}$, for EVV datasets (see Figure \ref{fig:pie_dataset_sample}). In addition to the sequence coordinates $\boldsymbol{X}$, $T_{obs}$ input frames/images are also available, denoted as $\boldsymbol{I} = \{I_{t} \ | \ t \in \left(1,2, \ldots ,T_{obs}\right)\}$.
The goal of ASTRA is to output deterministic or multi-modal trajectories of the pedestrian. In the deterministic setting, the problem consists of predicting the output prediction coordinates of the $A$ agents in the subsequent $T_{pred}$ future frames, formally $\boldsymbol{\hat{Y}} = \{\hat{Y}^{a}_{t} \ | \ t \in \left(1,2, \ldots ,T_{pred}\right)$$; \ a \in \left(1,2,\ldots,A\right)\}$ where $\hat{Y}^{a}_{t}$ denotes the coordinates of the agent $a$ at a future time $t$, and the corresponding ground truth being $\boldsymbol{Y} = \{Y^{a}_{t} \ | \ t \in \left(1,2, \ldots ,T_{pred}\right)$$; \ a \in \left(1,2,\ldots,A\right)\}$. To output multimodal trajectories we need to learn a generative model, denoted by $p_{\theta}(\mathcal{Y} | \boldsymbol{X}, \boldsymbol{I})$ which is parameterized by $\theta$ and given $\boldsymbol{X}$ and $\boldsymbol{I}$, outputs $K$ predicted trajectories denoted by $\mathcal{Y}=\{\boldsymbol{\hat{Y}^{(1)}}, \boldsymbol{\hat{Y}^{(2)}}, \ldots, $ $\boldsymbol{\hat{Y}^{(K)}}\}$. 

\subsection{Components}
\begin{figure*}[ht!]
\centering
  \includegraphics[width=0.6\linewidth]{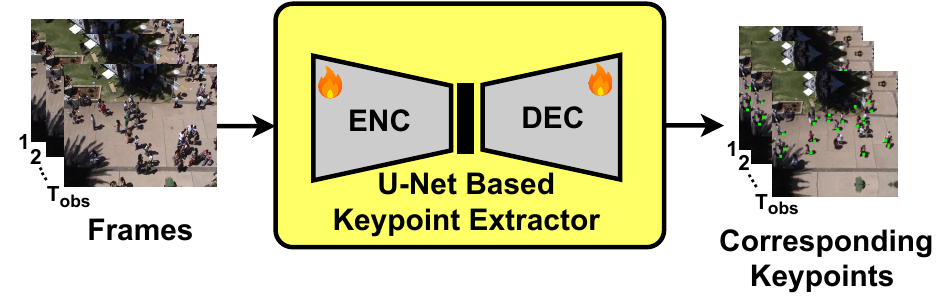} % Adjust the width here
  \caption{\small Pretraining U-Net based keypoint extractor.}
  \label{fig:pretraining}
\end{figure*}

The encoder part of our model consists of two main components: A scene-aware component and an agent-aware component. While the former is dedicated to encoding the scene, and encapsulating the contextual details, the latter focuses on encoding the spatial, temporal, and social dimensions of the agents, as shown in Figure \ref{fig:model}. The output from these two components is aggregated before being decoded to generate single or multiple predicted future trajectories for deterministic/multi-modal predictions, respectively. 
In order to learn essential information about the scene's spatial layout and the positional dynamics of agents within it, the U-Net \cite{ronneberger2015u} is pre-trained using the method detailed in \cite{ribera2019locating} which utilizes a specialized loss function, Weighted Hausdorff Distance to learn a latent representation of the scene context (Figure \ref{fig:pretraining}). More sophisticated schemes to generate the scene representation, like transformer-based architectures, and fusing social representations via gated cross-attention can also be considered but we leave exploring possibly more effective and sophisticated architecture designs as future work.
The Grad-CAM visualizations (Figures \ref{fig:eth_gradcam} and \ref{fig:pie_gradcam}) highlight this capability, showing that the pre-trained model pays attention to regions with a high likelihood of pedestrian activity. 
The U-Net-based keypoint extractor is frozen after the pretraining when used in ASTRA model architecture.

\subsubsection{Temporal Encodings}
To guide the network to model the temporal dimension, we add temporal encodings to the agents and scene to capture the temporal dependencies within the sequence of pedestrian trajectories from past frames, adopting a time encoder akin to the positional encoding found in the original Transformer architecture \cite{transformers}. 

\begin{equation} \label{eq:temporal_phi}
    \Phi_{\text{Temporal}}(t, i) = 
    \begin{cases}
    \sin\left(\frac{t}{10000^{2i/d}}\right) & \text{if } i \text{ is even}\\
    \cos\left(\frac{t}{10000^{2i/d}}\right) & \text{if } i \text{ is odd}
    \end{cases}
\end{equation}
where \(t\) is the time step, \(i\) is the dimension, and \(d\) is the dimensionality of the model.

\subsubsection{Scene-aware embeddings.}

\begin{figure*}
    \centering
    \begin{subfigure}{0.32\textwidth}
        \centering
        \includegraphics[height=4cm,width=\columnwidth]{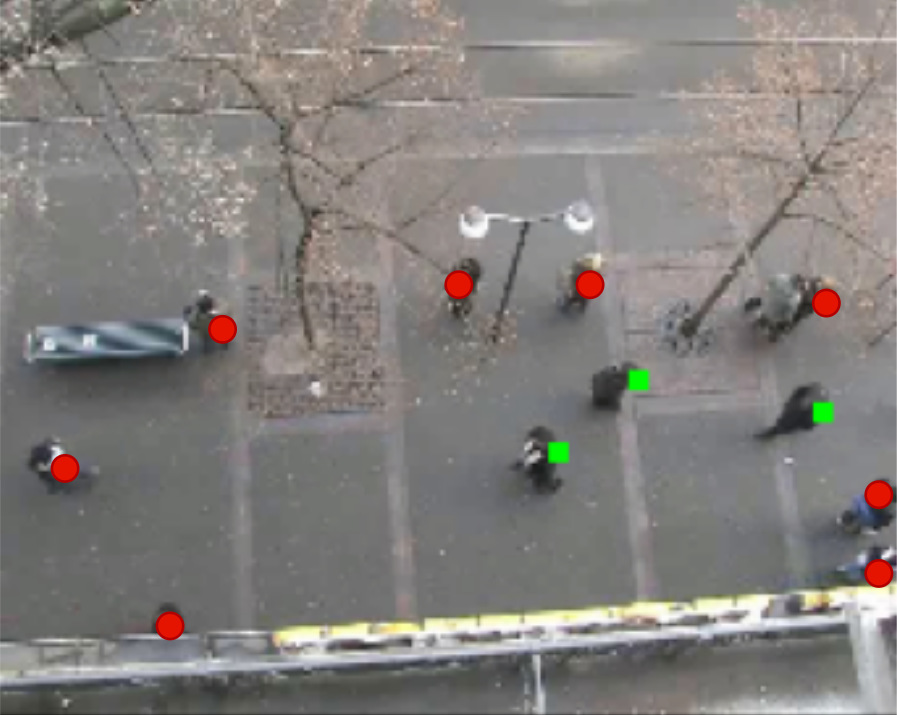}
        \caption{Original Image}
        \label{fig:original}
    \end{subfigure}
    % \hspace{0.1mm}
    \begin{subfigure}{0.32\textwidth}
        \centering
        \includegraphics[height=4cm,width=\columnwidth]{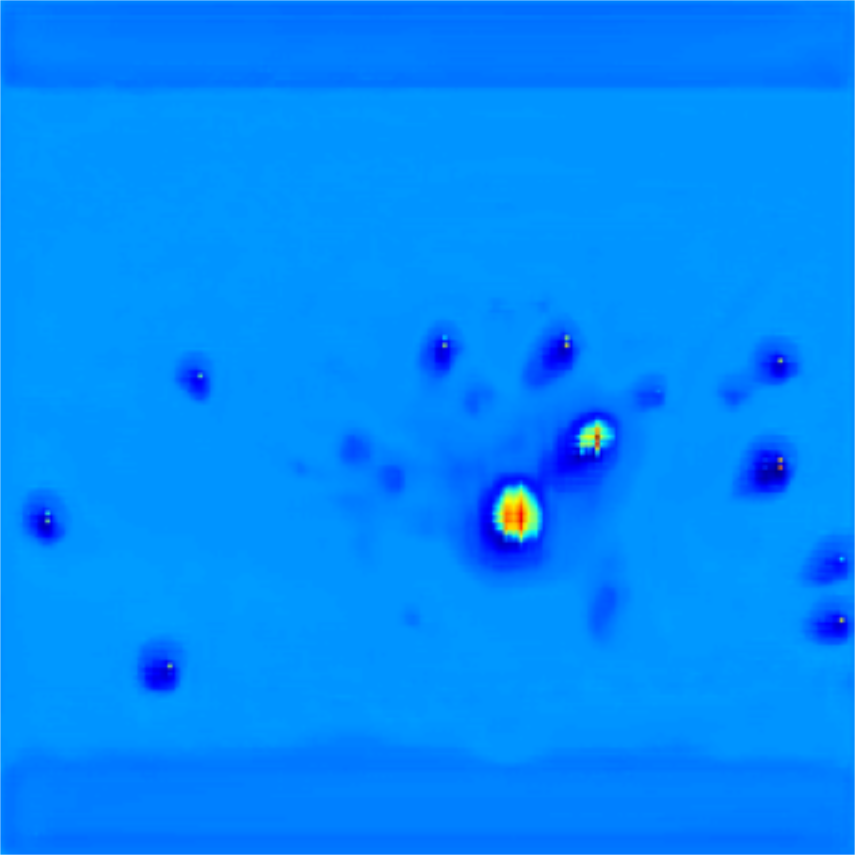}
        \caption{Grad-CAM Image}
        \label{fig:grad_img}
    \end{subfigure}
    % \hspace{0.1mm}
    \begin{subfigure}{0.32\textwidth}
        \centering
        \includegraphics[height=4cm,width=\columnwidth]{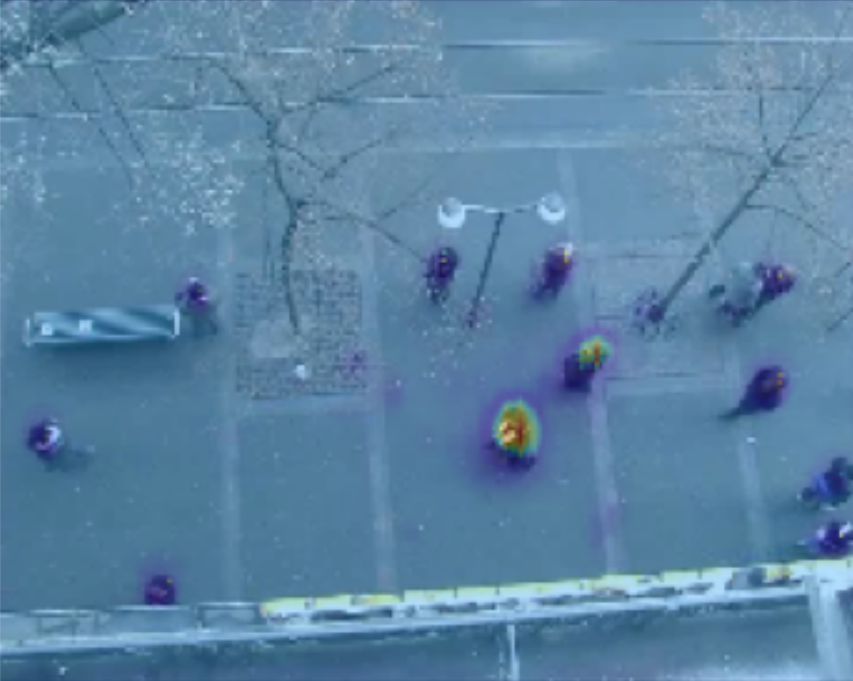}
        \caption{Overlayed Grad-CAM}
    \end{subfigure}
    
    \caption{\textbf{Grad-CAM visualizations}: In (a), the red circle indicates unlabelled pedestrians, while the green square highlights labelled pedestrians. In (b), the U-Net-based keypoint extractor focuses on unlabelled pedestrians as well, thereby capturing scene context from them too.}
    \label{fig:grad_cam_sup}
\end{figure*}

A latent representation of pedestrian characteristics is obtained using a pre-trained U-Net encoder (Figure \ref{fig:model}); this latent vector can include some crucial characteristics like spatial groupings and interactions with the environment. This step is crucial as the U-Net extractor possesses the proficiency to discern both labelled and unlabelled pedestrians, depicted in green and red, respectively in Figure \ref{fig:original}.

\subsubsection{Scene-aware transformer encoder.}
The latent representation of all frames ($\Psi_{\text{Scene}}$) is treated as input tokens to the scene-aware single-layer transformer encoder ($T_{\text{Scene-aware}}$), which in turn generates scene-aware embeddings ($\Phi_{\text{Scene}}$) for each frame. The single-layer transformer encoder architectural choice significantly contributes to the lightweight nature of our model. The resulting scene embedding is:

\begin{align}
\Psi_{\text{Scene}} &= \Gamma_{\text{Scene}}\left(\Upsilon_{\text{Encoder}}(\boldsymbol{I})\right) \\
\Phi_{\text{Scene}} &= T_{\text{Scene-aware}}\left([\Psi_{\text{Scene}}; \Phi_{\text{Temporal}}]\right)
\end{align}
where past input frame images projected using a Multi-Layer Perceptron (MLP) layer ($\Gamma_{\text{Scene}}$), and \(\Upsilon_{\text{Encoder}}(.)\) denotes the encoder part of the U-Net. 

The temporal encoding $\Phi_{\text{Temporal}}$, crucial for capturing the temporal dynamics within the observed frames, adopts the design of the traditional positional encoding \cite{transformers} and follows the work of \cite{AgentFormer}. 

\subsubsection{Agent-aware embeddings.}
The second component is dedicated to encoding the different dimensions of each agent for all agents in the scene.

The spatial coordinates $\mathbf{X}$ of each agent are linearly projected to a latent space using an MLP layer ($\Gamma_{\text{Spatial}}$) to get spatial encoding ($\Phi_{\text{Spatial}}$) 

\begin{equation}
    \Phi_{\text{Spatial}} = \Gamma_{\text{Spatial}}(\boldsymbol{X})
\end{equation}

Spatial embeddings are not enough to capture the full information about the agents. If two agents in two frames have the same location in the image, their spatial embedding will be the same. Hence, temporal encoding (equation in Supplementary material) is also included to distinguish them. 

Having the spatial and temporal dimensions of the agents is still not enough to understand their interaction in the scene. To capture the social dimension in this multi-agent environment, we generate a fully connected undirected graph between agents, in which the nodes are the agents' locations, and the edges between agents are the reciprocal of the distance. Consequently, the closer the agents are to each other, the stronger the link between them. Formally, we represent the social dimension using a graph $G = (V, E)$, where $V$ is the set of agents and $E$ is the collection of edges, with weights
\vspace{-2mm}
\begin{equation}
e_{ij} = \frac{1}{d(v_i, v_j)}
\end{equation}
\iffalse
\[ G = (V, E), \text{in which, $G$ is the graph} \]
\[ |V| \text{ set of nodes/agents locations} \]
\[ |E| \text{ set of edges, in which, }e_{ij} = \frac{1}{d(v_i, v_j)}, \]
\fi
where \( d(v_i, v_j) \) is the distance between agents \( v_i \) and \( v_j \).

Subsequently, Random Walk Positional Encodings (RWPEs) \cite{dwivedi2021graph} is used to capture the structural relationships between nodes in the graph, such as their proximity to each other and the number of paths between them. These RWPEs are further projected to latent space using a separate MLP ($\Gamma_{\text{Social}}$) to get social encodings ($\Phi_{\text{Social}}$), mathematically:
\begin{equation}
\Phi_{\text{Social}} = \Gamma_{\text{Social}}\left(\text{RWPE}(\mathbf{G})\right)
\end{equation}

This preserves the graph structure of the agents while making the transformer encoder graph-aware. Furthermore, our method empowers the network to determine the significance of each agent relative to others autonomously.%, in contrast to the approach in \cite{AgentFormer} where human-imposed biases influence the model by prescribing differentiated attention to an agent's own past compared to that of others. This autonomous evaluation by the network eliminates potential biases that may arise from predefined human inputs.

While we do not claim to be the first to use transformers or graphs, we claim that we are the first to integrate RWPE directly into transformer tokens, creating a graph-aware transformer in the context of trajectory prediction, which has proven successful, outperforming the state-of-the-art (SOTA).

\subsubsection{Agent-aware transformer encoder.}
After calculating the spatial, temporal and social representations for each agent, our model concatenates them. This concatenated vector is then fed into an agent-aware single-layer transformer encoder ($T_\text{Agent-aware}$) that generates agent-aware embedding (\(\Phi_{\text{Agents}}\)).

\vspace{-1.5mm}
{\small
\begin{equation}
\Phi_{\text{Agents}} = T_{\text{Agent-aware}}([\Phi_{\text{Spatial}};  \Phi_{\text{Temporal}}; \Phi_{\text{Social}}])
\end{equation}}
\vspace{-4mm}
\subsubsection{Decoder.}
To generate multiple stochastic trajectories, we learn a generative model,   $p_{\theta}(\mathcal{Y} | \boldsymbol{X}, \boldsymbol{I})$ for which we adopted CVAE(Conditional Variational Auto Encoder). We train CVAE to learn the inherent distribution of future target trajectories conditioned on observed past trajectories, by utilizing a latent variable $Z$. CVAE consists of three components - prior network (\(p_{\theta}(Z|X,I)\)), recognition network (\(q_{\phi}(Z | Y,X,I)\)) and generation network(\(g_{\nu}({\Tilde{Y}} | Z)\)), parameterized by $\theta$, $\phi$ and $\nu$ respectively. Here $\Tilde{Y}$ is the output of the generation network and is the latent representation of the future trajectories. To generate future trajectories, we pass $\Tilde{Y}$ to the MLP decoder ($\Gamma_{\text{Decoder}}$).

For deterministic predictions, CVAE is skipped and the outputs of both the scene transformer (\(\Phi_{\text{Scene}}\)) and the agents' transformer (\(\Phi_{\text{Agents}}\)) are concatenated and directly passed through an MLP decoder ($\Gamma_{\text{Decoder}}$) to produce future trajectories ($\hat{Y}$) of the agents in the future frames as shown in Figure \ref{fig:model}, namely: 
\begin{align}
\mathbf{\hat{Y}} = \Gamma_{\text{Decoder}}([\Phi_{\text{Scene}}; \Phi_{\text{Agents}}])
\label{eqn:deter}
\end{align}

\subsubsection{Weighted trajectory loss function.}

We introduce a weighted-penalty strategy that can be applied to common loss functions used in trajectory prediction such as MSE and Smooth L1 Loss. The application of this strategy is through a dynamic penalty function \( w(t) \), designed to escalate or de-escalate the significance of prediction errors as one moves further into the future. The definition of the weighted loss function is given by:
\begin{equation}
L_{\text{weighted}}(\hat{Y}, Y) = \sum_{t=1}^{T_{pred}} w(t) \cdot L(\hat{Y}_t, Y_t),
\end{equation}
where \( \hat{Y} \) and \( Y \) are the predicted and actual trajectories respectively, \( T_{pred} \) denotes the number of prediction timesteps, \( w(t) \) represents the dynamic weighting function at time \( t \), and \( L(\hat{Y}_t, Y_t) \) is the predefined loss function (e.g., MSE or SmoothL1 Loss) applied to the predicted and true positions at each time step \( t \).

In time series data, as we move further into the future relative to the last observed data, the drift in predictions tends to increase, leading to higher errors. Motivated by this intuition, we initially penalized the predictions using linear and quadratic loss functions. These approaches showed improvements in overall prediction accuracy.
However, upon closer analysis of the results, we observed that in some trajectories, there was a noticeable offset in the earlier parts of the predictions. To address this issue, we experimented with a parabolic weighting function for the penalty. Empirically, this approach outperformed the linear and quadratic strategies, yielding the most balanced and accurate predictions across the trajectories.

The weight function \( w(t) \) is designed to be versatile, accommodating a spectrum of mathematical formulations that align with the specific needs of the predictive model. It is generically defined as:
\begin{equation}
w(t) = f(t, T_{\text{pred}}, \alpha, \beta),
\end{equation}
where \( \alpha \) and \( \beta \) are parameters that establish the bounds of the weighting function, and \( f \) is an adaptable function that governs the progression of weights at each timestep \( t \).

In particular, the function \( w(t) \) may be selected from various mathematical forms, such as linear, parabolic, or quadratic, which are discussed in greater detail within the supplementary materials. The choice of function enables the model to adjust the penalty progression in alignment with the anticipated prediction challenge at each timestep. 

To generate multi-modal trajectories, the entire ASTRA model is optimized using Equation \ref{eqn:loss_multi}. The prior distribution (\(p_{\theta}(z|X,I)\)) is parameterized by \(\mathcal{N}(\mu^{p}_{z}, (\sigma^{p}_{z})^2)\). The approximate posterior distribution (\(q_{\phi}(z | Y,X,I)\)) is parameterized by \(\mathcal{N}(\mu^{q}_{z}, (\sigma^{q}_{z})^2)\), where  $\mu^{p}_{z}$ and $(\sigma^{p}_{z})^2$ represent the mean and variance of the prior distribution and $\mu^{q}_{z}$ and $(\sigma^{q}_{z})^2$ represent the mean and variance of the posterior distribution.
During Training, we sample latent variable($z$) from the recognition network (posterior distribution) and fed it to the generation network, whereas during testing we sample $z$ from the prior network (prior distribution). We use the reparameterization trick to sample $z$ through the mean and variance pairs of $(\mu^p_{z},(\sigma^p_{z})^2)$ and $(\mu^q_{z},(\sigma^q_{z})^2)$, respectively. KL divergence term help in minimizing the difference between the distribution of latent variable(z) of prior and recognition network. During inference, $K$ samples are drawn from the learned distribution and decoded to future trajectories. 

% The KL divergence term ensures that the prior network implicitly learns the dependency between $\boldsymbol{Y}$ and $\boldsymbol{X}$.

\begin{equation}
  \begin{aligned}
    \mathcal{L}_{\text{final}} &= L_{\text{weighted}}(\boldsymbol{\hat{Y}}, \boldsymbol{Y}) + 
    D_{KL}(q_{\phi}(Z | Y,X,I)||p_{\theta}(Z|X,I))
  \end{aligned}
  \label{eqn:loss_multi_1}
\end{equation}

\begin{equation}
  \begin{aligned}
    \mathcal{L}_{\text{final}} &= L_{\text{weighted}}(\boldsymbol{\hat{Y}}, \boldsymbol{Y}) + D_{KL}(\mathcal{N}(\mu_{z_{q}},\sigma_{z_{q}})|| \mathcal{N}(\mu_{z_{p}},\sigma_{z_{p}}))
  \end{aligned}
  \label{eqn:loss_multi}
\end{equation}

 % \Sachin{During inference, $K$ samples are drawn from the learned distribution and decoded to future trajectories.} are used to approximate the true distribution of the latent variables, enabling the generation of multiple plausible trajectories. The pair $(\mu_{z_{p}},\sigma_{z_{p}})$ is contrasted with $(\mu_{z_{q}},\sigma_{z_{q}})$, which parameterizes the ground truth distribution \(q_{\phi}(z_q | x, y)\), helping to guide the model towards accurate predictions through the KL divergence term. For deterministic trajectories loss, KL divergence term is dropped from the equation \ref{eqn:loss_multi}. More details are discussed in the supplementary. 

\section{Experiments}
\label{sec:exp}
\subsection{Datasets and Evaluation Protocol}
For a comprehensive evaluation, we benchmarked our model on three trajectory prediction datasets; namely, ETH \cite{eth2009}, UCY \cite{ucy2007}, and PIE dataset \cite{piedataset}. \textbf{ETH} and \textbf{UCY} offer a bird's-eye view of pedestrian dynamics in urban settings, including five datasets with 1,536 pedestrians across four scenes. For evaluation, we used their standard protocol; leave-one-out strategy, observing eight time steps (3.2s) and predicting the following 12 steps (4.8s).

In contrast, %to the bird's-eye view of the ETH-UCY dataset, 
the \textbf{PIE dataset} provides an Ego-vehicle perspective, containing over 6 hours of ego-centric driving footage, along with bounding box annotations for traffic objects, action labels for pedestrians, and ego-vehicle sensor information \cite{piedataset}. A total of 1,842 pedestrian samples are considered with the following split: Training(50\%), Validation(40\%) and Testing(10\%)\cite{piedataset}. Model performance is evaluated based on a shorter observational window of 0.5 seconds and a prediction window of 1 second, providing insights into the model's capability in rapidly evolving traffic scenarios\cite{rasouli2023pedformer}.

\subsection{Evaluation Metrics}
We used the standard evaluation metrics of ADE and FDE for ETH-UCY deterministic settings and minADE and minFDE for their stochastic settings. ADE, FDE, CADE, CFDE, ARB and FRB for PIE dataset, the supplementary material explains these metrics. 

\subsection{Setting up the experiments}
We trained the model on the ETH-UCY and PIE datasets using the AdamW optimizer with a weight decay of \(5 \times 10^{-4}\) for 200 epochs. The initial learning rate was set to \(1 \times 10^{-3}\), and a cosine annealing scheduler was employed. Training was conducted on a NVIDIA DGX A100 machine, equipped with 8 GPUs, each having 80 GB of memory.

\subsection{FLOPS}
For AgentFormer, the model has approximately 3.084 GFLOPs, whereas ASTRA is comparatively lighter with 1.7 MFLOPs (including U-Net and CVAE), highlighting ASTRA's computational efficiency. With pretrained U-net the model has 839 KFLOPs and without including the CVAE, model has 16 KFLOPs. 

% \begin{table}[ht]
% \centering
% \caption{Comparison of FLOPs and Parameters for AgentFormer and ASTRA}
% \begin{tabular}{lcc}
% \toprule
% \textbf{Model} & \textbf{FLOPs (MegaFLOPs)} & \textbf{Parameters (Millions)} \\
% \midrule
% % AgentFormer & 3084 & 23.4 \\
% ASTRA       & 1.7 & 0 \\
% ASTRA(pretrained U-net)       & 0.839 & 0 \\
% ASTRA(pretrained U-net, without CVAE)       & 0.016 & 0 \\

% \bottomrule
% \end{tabular}
% \label{tab:comparison_flops_parameters}
% \end{table}

\subsection{Discussing the Results}
\subsubsection{Quantitative Results.}

\begin{table*}[ht!]
    \centering    \caption{\textbf{Deterministic Results}: ADE/FDE results for ETH-UCY baselines. Best in \textbf{bold}, second best \underline{underlined}.}
    \label{tab:eth_Baseline}
    % \resizebox{\columnwidth}{!}{
    \small
    \begin{tabular*}{1.02\textwidth}{@{}lcccccc@{}}
    \noalign{\hrule height 1pt}
    % \multicolumn{1}{c}{\textbf{}} & \multicolumn{6}{c}{\textbf{ADE/FDE}} \\
    % \cline{2-7}
    \textbf{Model} & \textbf{ETH} & \textbf{Hotel} & \textbf{Univ} & \textbf{Zara1} & \textbf{Zara2} & \textbf{Average} \\
    \hline
    Linear & 1.33/2.94 & 0.39/0.72 & 0.82/1.59 & 0.62/1.21 & 0.77/1.48 & 0.79/1.59 \\
    S-LSTM\cite{S-LSTM} & 1.09/2.35 & 0.79/1.76 & 0.67/1.40 & 0.47/1.00 & 0.56/1.17 & 0.72/1.54 \\
    S-Attention\cite{sattention} & 1.39/2.39 & 2.51/2.91 & 1.25/2.54 & 1.01/2.17 & 0.88/1.75 & 1.41/2.35 \\
    SGAN-ind\cite{SGAN} & 1.13/2.21 & 1.01/2.18 & 0.60/1.28 & 0.42/0.91 & 0.52/1.11 & 0.74/1.54 \\
    Traj++\cite{Traj++} & 1.02/2.00 & 0.33/0.62 & \underline{0.53}/1.19 & 0.44/0.99 & 0.32/0.73 & 0.53/1.11 \\
    TransF\cite{transf} & 1.03/2.10 & 0.36/0.71 & \underline{0.53}/1.32 & 0.44/1.00 & 0.34/0.76 & 0.54/1.17 \\
    MemoNet\cite{memonet} & 1.00/2.08 & 0.35/0.67 & 0.55/1.19 & 0.46/1.00 & 0.37/0.82 & 0.55/1.15 \\
    SGNet\cite{sgnet} & \underline{0.81}/\underline{1.60} & 0.41/0.87 & 0.58/1.24 & \underline{0.37}/\underline{0.79} & 0.31/0.68 & 0.56/1.04 \\
    EqMotion\cite{EqMotion} & 0.96/1.92 &	\underline{0.30}/\underline{0.58} & \textbf{0.50}/\underline{1.10} &	0.39/0.86 &	\underline{0.30}/\underline{0.68} &	\underline{0.49}/\underline{1.03} \\ 

    % LMTraj-SUP\cite{Lmtraj} & \underline{0.65}/\underline{1.04} & \textbf{0.26}/\textbf{0.46} & 0.57/1.16 & 0.51/1.01 & 0.38/0.74 & 0.48/\underline{0.88} \\
    
    % SingularTrajectory\cite{singulartrajectory} & 0.72/1.23 & \underline{0.27}/\underline{0.50} & 0.57/1.12 & 0.44/0.93 & 0.35/0.73 & \underline{0.47}/0.90 \\

\hline
    \vspace{0.5mm}
    
    ASTRA (Ours) & \textbf{0.47}/\textbf{0.82}  & \textbf{0.29}/\textbf{0.56}   & 0.55\textbf{/1.00}  & \textbf{0.34}/\textbf{0.71}  & \textbf{0.24}/\textbf{0.41} & \textbf{0.38}/\textbf{0.70} \\ 
         \noalign{\hrule height 1pt}
    \end{tabular*}
    % }
\end{table*}

\begin{table*}[ht!]
\centering

\caption{\textbf{Stochastic Results}: minADE$_{20}$ and minFDE$_{20}$ results for ETH-UCY baselines. Best in \textbf{bold}, second best \underline{underlined}. NP- means unpenalised.}
\small
\label{tab:model_performance}
\begin{tabular*}{1.025\textwidth}{@{\extracolsep{\fill}}lcccccc}
\noalign{\hrule height 1pt}
\textbf{Model} & \textbf{ETH} & \textbf{Hotel} & \textbf{Univ} & \textbf{Zara1} & \textbf{Zara2} & \textbf{Average} \\ \hline
Social-GAN \cite{SGAN}& 0.87/1.62 & 0.67/1.37 & 0.76/1.52 & 0.35/0.68 & 0.42/0.84 & 0.61/1.21 \\
NMMP  \cite{nmmp}& 0.61/1.08 & 0.33/0.63 & 0.52/1.11 & 0.32/0.66 & 0.43/0.85 & 0.41/0.82 \\
STAR \cite{star} & 0.36/0.65 & 0.17/0.36 & 0.31/0.62 & 0.29/0.52 & 0.22/0.46 & 0.26/0.53 \\
PECNet  \cite{pecnet}      & 0.54/0.87 & 0.18/0.24 & 0.35/0.60 & 0.22/0.39 & 0.17/0.30 & 0.29/0.48 \\
Trajectron++ \cite{Traj++} & 0.61/1.02 & 0.19/0.28 & 0.30/0.54 & 0.24/0.42 & 0.18/0.32 & 0.30/0.51 \\
BiTrap-NP  \cite{yao2021bitrap}    & 0.55/0.95 & 0.17/0.28 & 0.25/0.47 & 0.22/0.44 & 0.16/0.33 & 0.27/0.49 \\
MemoNet \cite{memonet}      & 0.40/0.61 & \textbf{0.11}/\textbf{0.17} & 0.24/\underline{0.43} & 0.18/0.32 & 0.14/0.24 & \underline{0.21}/0.35 \\
GroupNet \cite{Xu_2022_CVPR}  & 0.40/0.76 & \underline{0.12}/\underline{0.18} & \textbf{0.22}/\textbf{0.41} & \underline{0.17}/0.31 & \textbf{0.12}/0.24 & \underline{0.21}/0.38 \\
SGNet \cite{sgnet}& 0.47/0.77 & 0.20/0.38 & 0.33/0.62 & 0.18/0.32 & 0.15/0.28 & 0.27/0.47 \\
MID   \cite{Gu_2022_CVPR} & 0.46/0.73 & 0.15/0.25 & 0.26/0.49 & 0.21/0.39 & 0.17/0.33 & 0.25/0.44 \\
Agentformer  \cite{AgentFormer} & 0.45/0.75 & 0.14/0.22 & 0.25/0.45 & 0.18/0.30 & 0.14/0.24 & 0.23/0.39 \\
EqMotion  \cite{EqMotion} & 0.40/0.61 & \underline{0.12}/\underline{0.18} & \underline{0.23}/\underline{0.43} & 0.18/0.32 & \underline{0.13}/0.23 & \underline{0.21}/0.35 \\ 
Leapfrog  \cite{leapfrog} & 0.39/0.58 & \textbf{0.11}/\textbf{0.17} & 0.26/\underline{0.43} & 0.18/\underline{0.26} & \underline{0.13}/0.22 & \underline{0.21}/0.33 \\

% \textcolor{blue}{LMTraj-SUP\cite{Lmtraj} & 0.41/0.51 & 0.12/0.16 & 0.22/0.34 & 0.20/0.32 & 0.17/0.27 & 0.22/0.32} \\
    
% \textcolor{blue}{E-V\textsuperscript{2}-Net-SC\cite{ev2} & 0.25/0.38
% & 0.12/0.14 & 0.20/0.34 & 0.18/0.29 & 0.13/0.22 & 0.17/0.27 }
%     \\ 
% \textcolor{blue}{SingularTrajectory\cite{singulartrajectory} & 0.35/0.42 & 0.13/0.19 & 0.25/0.44 & 0.19/0.32 & 0.15/0.25 & 0.21/0.32} \\
% \textcolor{blue}{MS-TIP\cite{mstip} & 0.39/0.57 & 0.13/0.22 & 0.24/0.40 & 0.20/0.34 & 0.17/0.29 & 0.22/0.36} \\

% \textcolor{blue}{HighGraph\cite{highgraph} & 0.33/0.56 & 0.13/0.21 & 0.23/0.47 & 0.19/0.33 & 0.15/0.25 & 0.21/0.36 }\\

\hline
ASTRA(Non Penalised) & 0.37/0.49     & 0.24/0.34     & 0.37/0.52     & 0.23/0.32     & 0.16/0.23     & 0.27/0.38 \\
ASTRA(Without Frame Encoding) & \underline{0.29}/\underline{0.39} & 0.18/ 0.29 & 0.29/ \underline{0.43} & \underline{0.17}/\underline{0.26} & 0.14/\underline{0.2} & \underline{0.21}/ \underline{0.31} \\
ASTRA(With Frame Encoding)        & \textbf{0.27}/\textbf{0.36}     & 0.17/0.25     & 0.28/\textbf{0.41}     & \textbf{0.15}/\textbf{0.23}     & \underline{0.13}/\textbf{0.16}     & \textbf{0.20}/\textbf{0.28} \\ 
\noalign{\hrule height 1pt}
\end{tabular*}
\end{table*}

For ETH-UCY, we compared our model results against several baselines. These comparisons are presented in Table~\ref{tab:eth_Baseline} and Table~\ref{tab:model_performance}, which contains results primarily sourced from the EqMotion (CVPR 2023) \cite{EqMotion} for deterministic predictions and LeapFrog (CVPR 2023) \cite{leapfrog} for stochastic predictions respectively. It is important to note that to provide a thorough comparative framework, we independently computed and included additional models \cite{sgnet,yao2021bitrap} to their respective tables as they were not originally part of the EqMotion or LeapFrog analysis. 
Our model significantly advances the state-of-the-art on ETH-UCY, outperforming the EqMotion \cite{EqMotion} by improving predictive accuracy by approximately 27\% on average for deterministic predictions as shown in Table \ref{tab:eth_Baseline} and approximately 10\% on average improvement over LeapFrog\cite{leapfrog} for stochastic predictions as shown in Table  \ref{tab:model_performance}, highlighting the efficacy of our approach in diverse scenarios.
We also highlight the effectiveness of utilizing frame encodings from U-Net, as demonstrated in Table \ref{tab:model_performance}.

Similarly, we benchmarked our model against various established models for the PIE dataset. The comparative analysis is summarized in Table \ref{tab:pie_baseline}, with the reference results taken from the PedFormer \cite{rasouli2023pedformer}, demonstrating an average improvement of 26\%. 

\textbf{Trainable Parameters and Computational Efficiency .} 

Regarding the computational side, ASTRA has seven times fewer trainable parameters than the existing SOTA model LeapFrog \cite{leapfrog} as shown in Figure \ref{fig:param-compare}. It is important to note that the parameter count reported for ASTRA in Figure \ref{fig:param-compare} includes the parameters from the U-Net, which is otherwise actually frozen during the trajectory prediction phase. The full ASTRA architecture with all the components included involves 1.56M parameters and 1.77M FLOPs, which is substantially more efficient than AgentFormer’s 6.78M parameters and 3.08G FLOPs. Even configurations of ASTRA without U-Net pretraining for deterministic predictions can be as compact as 13.52K parameters with minimal FLOPs (15.87K), indicating that we can flexibly adapt ASTRA’s complexity to meet strict resource constraints. A more detailed comparison for the same is given in \texttt{supplementary}.

\subsubsection{Qualitative results.}

\begin{figure}[h!]
    \centering
    \begin{subfigure}{0.45\textwidth}
        \centering
        \includegraphics[width=\columnwidth]{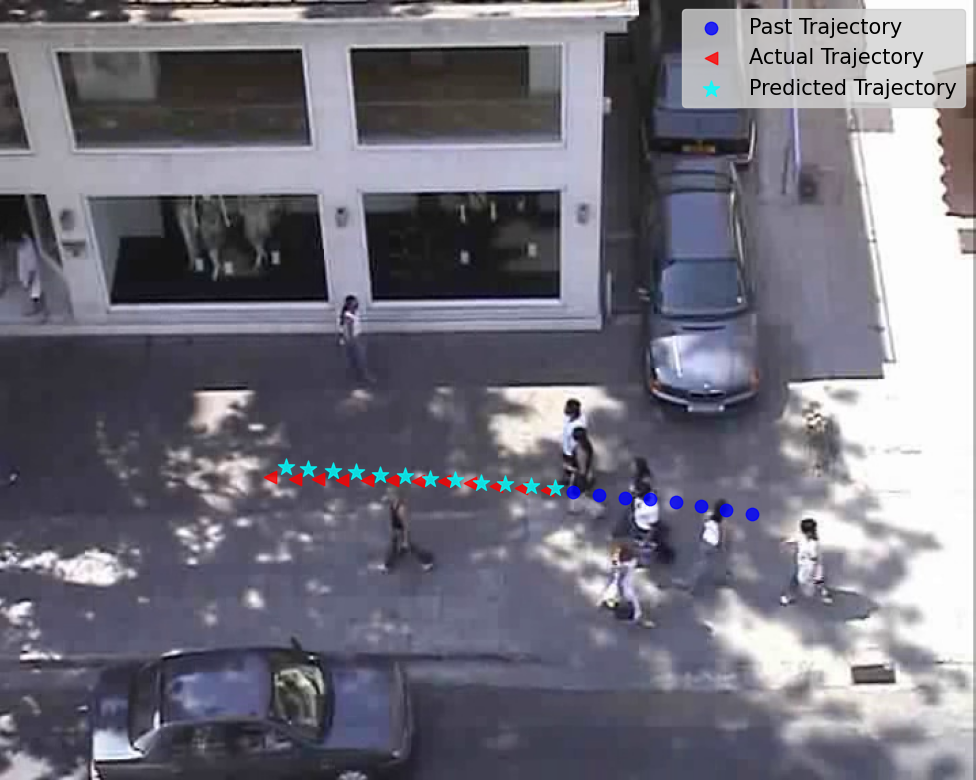}
        \caption{BEV}
    \end{subfigure}
    % \hspace{0.2mm}
    \begin{subfigure}{0.45\textwidth}
        \centering
        \includegraphics[height = 0.8\columnwidth,width=\columnwidth]{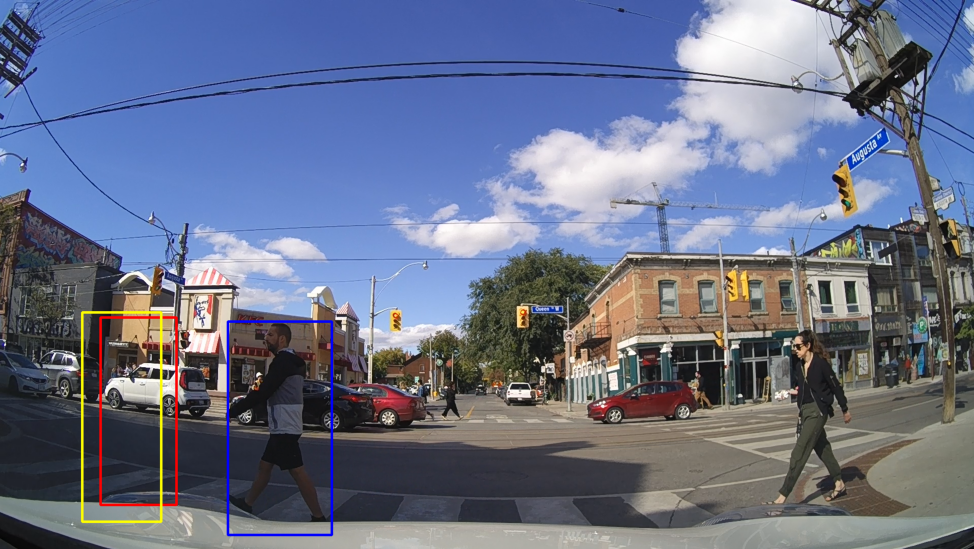}
        \caption{EVV}
    \end{subfigure}
    
    \caption{Sample images of the deterministic prediction from BEV datasets (a.) (ETH and UCY) and EVV dataset (b.) (PIE). The Red and Yellow bounding box indicates the ground-truth and predicted final position respectively and the Blue bounding box indicates the start position. }
    \label{fig:predict_datasets}
\end{figure}

\begin{figure*}[ht!]
    \centering
    \begin{minipage}{0.45\textwidth}
        \centering
        \includegraphics[width=\textwidth]{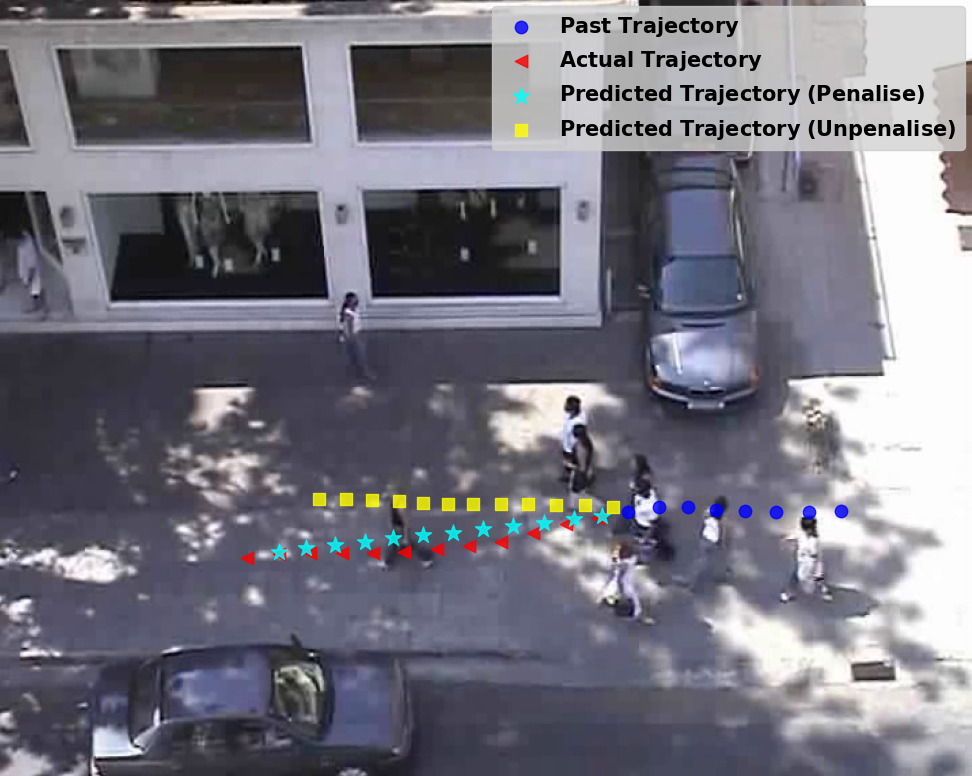}
        \caption{Visual comparison of penalised and unpenalised loss on ETH-UCY, showing the enhanced performance of the former.}
        \label{fig:qualitative_res_penalised}
    \end{minipage}
    \hfill
    \begin{minipage}{0.45\textwidth}
        \vspace{-0.42 cm}
        \centering
        \includegraphics[width=\textwidth]{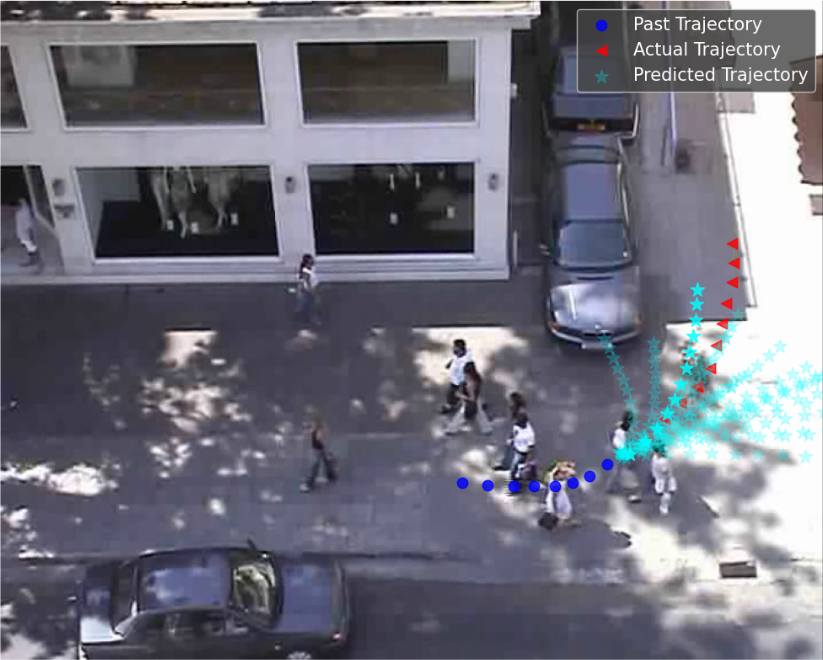}
        \caption{Sample images of the stochastic predictions from ETH-UCY Dataset.}
        \label{fig:stoch_results}
    \end{minipage}
\end{figure*}

We can clearly see the results of our prediction from Figure \ref{fig:predict_datasets} for deterministic predictions and Figure \ref{fig:stoch_results} for stochastic prediction. Figure \ref{fig:qualitative_res_penalised} exemplifies the proximity of our model's results to the ground truth, it also shows how using the weighted penalty strategy has yielded better results than the unpenalised one, highlighting the improved effectiveness of our strategy.
\vspace{-4mm}

\subsubsection{Ablation studies.} 
\begin{table}[h!]
\centering
\caption{\textbf{Ablation:} ADE/FDE \& minADE$_{20}$/minFDE$_{20}$ with variations in ASTRA's model components on UNIV dataset (where $\checkmark$: Component enabled, $\times$: Component disabled)}
\small % Adjust the font size
\setlength{\tabcolsep}{3.5pt} % Adjust the column separation
\vspace{-2mm}
\label{table:ablation}
\begin{tabular}{@{}ccccccc@{}}
\toprule
\textbf{Spatial} & \textbf{Temporal} & \textbf{Augmentation} & \textbf{Social} & \textbf{\begin{tabular}[c]{@{}c@{}}U-Net\\ Features\end{tabular}}& \textbf{ADE/FDE} & \textbf{\begin{tabular}[c]{@{}c@{}}\textbf{minADE$_{20}$}/\\ \textbf{minFDE$_{20}$}\end{tabular}}\\ \midrule
$ \checkmark $ & $ \times $ & $ \times $ & $ \times $ & $ \times $ & 1.05/1.66 & 0.43/0.63\\
$ \checkmark $ & $ \checkmark $ & $ \times $ & $\times$ & $ \times $ & 0.86/1.47 & 0.39/0.54\\
$ \checkmark $ & $\checkmark$ & $\checkmark$ & $\times$ & $ \times $ & 0.67/1.17 & 0.31/0.48\\
$\checkmark$ & $\checkmark$ & $\checkmark$ & $\checkmark$ & $ \times $ & 0.66/1.12 & 0.29/0.43\\ 
$\checkmark$ & $\checkmark$ & $\checkmark$ & $\checkmark$ & $ \checkmark $ & 0.55/1.00 & 0.28/0.41\\ 
\bottomrule
\end{tabular}
\end{table}

In the ablation study presented in Table \ref{table:ablation}, we evaluated the contribution of each component in our trajectory prediction model to ascertain their individual and collective impact on the performance metrics on the ETH-UCY (UNIV) dataset. Initially, the model incorporated only spatial information, which served as a baseline for subsequent enhancements. The sequential integration of temporal and social components yielded successive improvements, demonstrating their respective significance in capturing the dynamics of agent movement. The addition of the data augmentation technique (detailed in the supplementary material) further refined the model's performance, illustrating the value of varied training samples in enhancing generalization capabilities. Moreover, the incorporation of U-Net features contributed to a substantial leap forward, highlighting the importance of context-aware embeddings in accurately forecasting agent trajectories. This progression emphasizes the synergistic effect of combining heterogeneous data representations to capture the nuanced patterns of movement within a scene.

The ablation study also extended to the evaluation of loss functions, comparing the effects of penalised versus unpenalised approaches. Penalized loss functions, designed to focus the model's attention on more critical prediction horizons, proved to be more effective in refining the predictive accuracy, as outlined in Table \ref{table:loss} and in Table \ref{tab:model_performance} (ASTRA(NP)) for deterministic and stochastic setting respectively and the same can be observed in Figure \ref{fig:qualitative_res_penalised}. 

\begin{table}[h!]
\begin{minipage}[b]{0.45\linewidth}
\centering
\caption{Results for PIE dataset.}
\label{tab:pie_baseline}
\resizebox{1.20\linewidth}{!}{
\begin{tabular}{lcccc}
\noalign{\hrule height 1pt}
\textbf{Model} & \textbf{CADE} & \textbf{CFDE} & \textbf{ARB} & \textbf{FRB}  \\
\hline
FOL\cite{fol} & 73.87 & 164.53 & 78.16 & 143.69 \\
FPL\cite{fpl} & 56.66 & 132.23 & - & - \\
B-LSTM\cite{BLSTM} & 27.09 & 66.74 & 37.41 & 75.87 \\
PIE\textsubscript{traj}\cite{piedataset} & 21.82 & 53.63 & 27.16 & 55.39 \\
PIE\textsubscript{full}\cite{piedataset} & 19.50 & 45.27 & 24.40 & 49.09 \\
BiPed\cite{BiPed} & 15.21 & 35.03 & 19.62 & 39.12 \\
PedFormer\cite{rasouli2023pedformer} & \underline{13.08} & \underline{30.35} & \textbf{15.27} & \underline{32.79} \\ \hline
ASTRA(Ours) & \textbf{9.91}& \textbf{22.42} & \underline{18.32}  & \textbf{17.07}  \\
\noalign{\hrule height 1pt}
\end{tabular}
}
\end{minipage}%
\hfill
\begin{minipage}[b]{0.45\linewidth}
\centering
\caption{\textbf{Ablation:} ADE/FDE for penalised vs. unpenalised loss functions on UNIV dataset using SOTA ASTRA's configuration}
\label{table:loss}
{\small
\begin{tabular}{lcc}
\toprule
\textbf{Loss} & \textbf{Normal} & \textbf{Penalised}\\ \midrule
MSE &  0.58/1.13 &  0.57/1.00\\
SmoothL1 &  0.57/1.15 & 0.55/1.00\\ \bottomrule
\end{tabular}
}
\end{minipage}
\end{table}

\section{Conclusion}
\label{sec:conc}
We presented ASTRA, a model in the domain of pedestrian trajectory prediction, that outperforms the existing state-of-the-art models.
This advancement renders ASTRA particularly suitable for deployment on devices with limited processing capabilities, thereby broadening the applicability of high-accuracy trajectory prediction technologies.
ASTRA's adeptness in handling both BEV and EVV modalities further solidifies its applicability in diverse operational contexts. With the ability to produce deterministic and stochastic results, it enhances the predictive robustness and situational awareness of autonomous systems. Moving forward, we aim to extend the capabilities of the ASTRA model beyond pedestrian trajectory prediction to encompass a broader range of non-human agents. This expansion will involve adapting the model to understand and predict the movements of various entities within shared environments using more sophisticated architectural design choices to encode the scene and its fusion with social dimension. By broadening our focus, we hope to contribute to the development of truly comprehensive and adaptive systems capable of navigating the complexities of real-world interactions among a wide array of agents.

\clearpage
\bibliography{main}
\bibliographystyle{IEEEtran}

\clearpage
\appendix
\section{Conditional Variational Auto-Encoder Preliminaries}

ASTRA employs a Conditional Variational Autoencoder (CVAE) \cite{kingma2013auto} framework to address the inherent stochasticity of the prediction task, enabling the generation of $K$ distinct trajectories for each agent under consideration. Throughout the training phase, it aims to approximate the latent distribution $Z$ by deducing its mean and variance by utilising Multilayer Perceptrons (MLPs). Subsequently, by employing the divergence loss function (second term in Equation \ref{eq:lfinal}), the model pushes the learned distribution \(p_{\theta}(z_p | x)\), parameterized by $\theta$, to be as close as possible to the ground truth distribution \(q_{\Phi}(z_q | x, y)\), parameterized by $\Phi$. We use the reparameterization trick to present $z_p$ and $z_q$ through the mean and variance pairs of $(\mu_{z_{p}},\sigma_{z_{p}})$ and $(\mu_{z_{q}},\sigma_{z_{q}})$, respectively. After training, $K$ samples are drawn from the $Z$ distribution and decoded to form the final trajectories.

\section{Loss Function Formulation}
The loss for multi-modal trajectories is given in equations \eqref{eq:lweighted} and 
\eqref{eq:lfinal}. \\ $L_{\text{weighted}}(\mathcal{Y}_k, \boldsymbol{Y})$ is calculated similar to Equation \ref{eq:ldeter}.

\begin{align}
    % L_{\text{weighted}}(\boldsymbol{\hat{Y}}, \boldsymbol{Y}) &= L_{\text{weighted}}(\gamma, \boldsymbol{Y}) +  \min_{k = 1, \ldots, K} L_{\text{weighted}}(\mathcal{Y}_k, \boldsymbol{Y}) 
    L_{\text{weighted}}(\boldsymbol{\hat{Y}}, \boldsymbol{Y}) &= \min_{k = 1, \ldots, K} L_{\text{weighted}}(\mathcal{Y}_k, \boldsymbol{Y})  
    % = \sum_{t=1}^{T_{pred}} w(t) \cdot L(\hat{Y}_t, Y_t),
    \label{eq:lweighted} \\
    \mathcal{L}_{\text{final}} &= L_{\text{weighted}}(\boldsymbol{\hat{Y}}, \boldsymbol{Y}) + D_{KL}(\mathcal{N}(\mu_{z_{q}},\sigma_{z_{q}}) \ || \ \mathcal{N}(\mu_{z_{p}},\sigma_{z_{p}})) \label{eq:lfinal}
\end{align}

For deterministic predictions, the final loss is the same as the weighted loss:
\begin{equation}
\mathcal{L}_{\text{final}} = L_{\text{weighted}}(\hat{Y}, Y) = \sum_{t=1}^{T_{pred}} w(t) \cdot L(\hat{Y}_t, Y_t),
\label{eq:ldeter}
\end{equation}

where $w(t)$ represent the weighted penalty function (section \ref{weighted_penalty}) and $L(\hat{Y}_t, Y_t)$ is the predefined loss function: MSE or Smooth L1 loss (discussed below).

\textbf{Mean square error (MSE)}

\begin{equation}
\text{MSE} = \frac{1}{N} \sum_{i=1}^{N} (y_i - \hat{y}_i)^2
\end{equation}
where \( y_i \) and \( \hat{y}_i \) represent, the actual and predicted coordinates, respectively. MSE penalises larger trajectory prediction errors more heavily, ensuring model accuracy in critical scenarios. 

\textbf{Smooth L1 loss (SL1)}
\begin{equation}
\text{SL1}(y_i, \hat{y}_i) = \begin{cases} 
  0.5 \times (y_i - \hat{y}_i)^2 & \text{if } |y_i - \hat{y}_i| < 1 \\
  |y_i - \hat{y}_i| - 0.5 & \text{otherwise}.
\end{cases}
\end{equation}
Unlike MSE, SL1 effectively balances the treatment of small and large errors. This loss is also less sensitive to outliers, due to its combination of L1 and L2 loss properties.

\section{Weighted Penalty Functions}
\label{weighted_penalty}
\begin{figure}[!ht]
    \centering
    \includegraphics[height=0.4\textwidth]{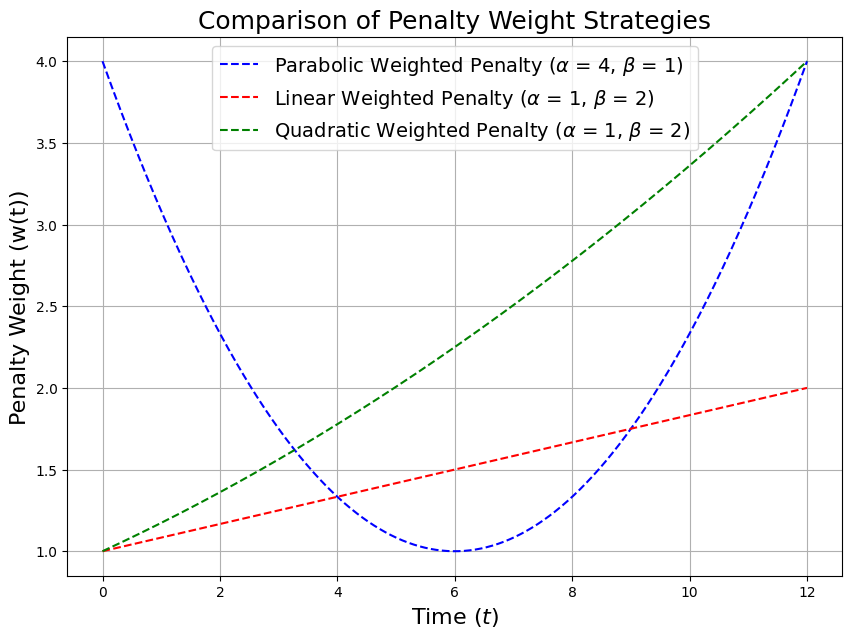}
    \caption{Comparison of various weighted penalty strategies} 
    \label{fig:penalties}
\end{figure}

In trajectory prediction, particularly for dynamic entities like pedestrians, vehicles, or other agents, the accuracy of predictions is paramount. The inherent challenge lies in managing the variability and uncertainty that escalates with longer prediction horizons. To address this, we implement different penalization strategies that adjust the model's emphasis to enhance reliability over extended prediction horizons.
Our ablation focuses on three distinct penalty strategies: Linear, Quadratic, and Parabolic. Table \ref{table:loss_p} presents a quantitative analysis comparing the three penalty strategies as applied to the ETH-UCY (UNIV) dataset using SL1 loss. It can be observed that the Parabolic penalty gives better results compared to other penalization strategies. Figure \ref{fig:penalties} compares the three weighted penalty strategies for a prediction window of 12 frames. Subsequent sections provide a detailed explanation for each of the penalty strategies.

\subsection{Linear Weighted Penalty}
The Linear Weighted Penalty employs a weight function, \(w(t)\), that linearly increases from a start weight (\(\alpha\)) to an end weight (\(\beta\)), over the prediction period. This approach aims to progressively increase the penalty for prediction inaccuracies, particularly toward the latter part of the prediction horizon.

The weight function $w(t)$ is defined as:
\begin{equation}
w(t) = \alpha + \frac{t}{T_{pred}} \cdot (\beta - \alpha),
\end{equation}
where $\alpha$ and $\beta$ are the weights assigned to the initial and final predicted time steps, respectively.
% defined start and end weights for the penalty.

\subsection{Quadratic Weighted Penalty}
The quadratic weighted penalty strategy intensifies the penalty in a quadratic manner as the difference between the prediction time and the past frames increases. This approach is more aggressive than the linear strategy, applying an exponentially increasing weight to errors in later prediction frames. The weight function \( w(t) \) in this case is defined as:

\begin{equation}
    w(t) = \left( \alpha + \frac{t}{T_{\text{pred}}} \cdot (\beta - \alpha) \right)^2
\end{equation}

\subsection{Parabolic Weighted Penalty}
The Parabolic Weighted Penalty assigns the maximum weight, \(\alpha\), to both the initial and final predicted time steps, highlighting their significance. Meanwhile, the minimum weight, \(\beta\) (\(\beta<\alpha\)), is allocated to the midpoint of the prediction interval. This distribution forms a parabolic trajectory (shown in Figure \ref{fig:penalties}) of weights across the prediction period, as defined by:

\begin{equation}
w(t) = (\alpha - \beta) \cdot \left(2 \cdot \frac{t}{T_{pred}} - 1\right)^2 + \beta,
\end{equation}

\begin{figure*}
    \centering
    % First row of images
    \begin{subfigure}{0.475\textwidth}
        \centering
        \includegraphics[height=4.5cm,width=\columnwidth]{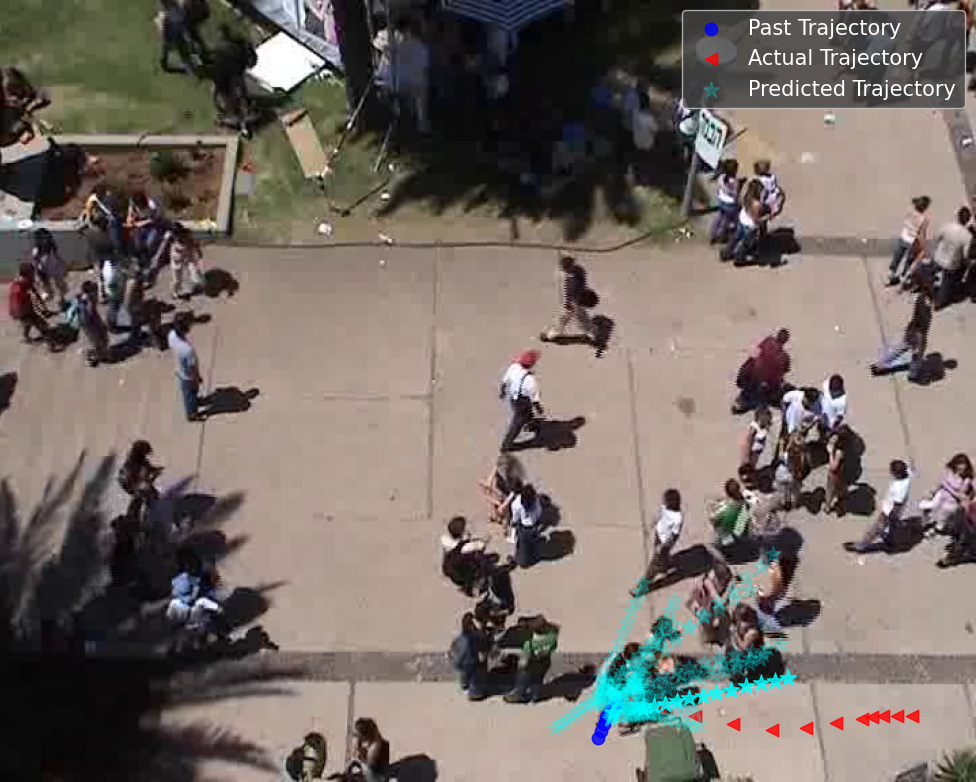}
        \caption{Univ (No Penalty)}
        \label{subfig:univ_no_pen1}
    \end{subfigure}
    \hspace{0.1mm}
    \begin{subfigure}{0.475\textwidth}
        \centering
        \includegraphics[height=4.5cm,width=\columnwidth]{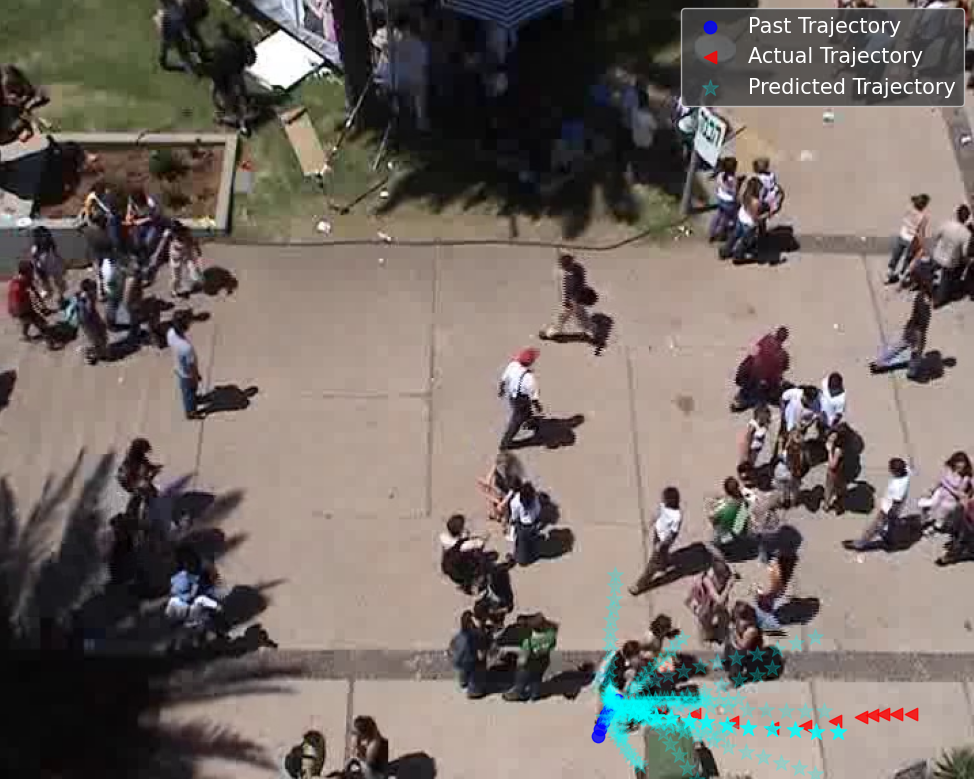}
        \caption{Univ (Penalty)}
        \label{subfig:univ_pen1}
    \end{subfigure}
    
    \vspace{5mm} % Space between rows

    \begin{subfigure}{0.475\textwidth}
        \centering
        \includegraphics[height=4.5cm,width=\columnwidth]{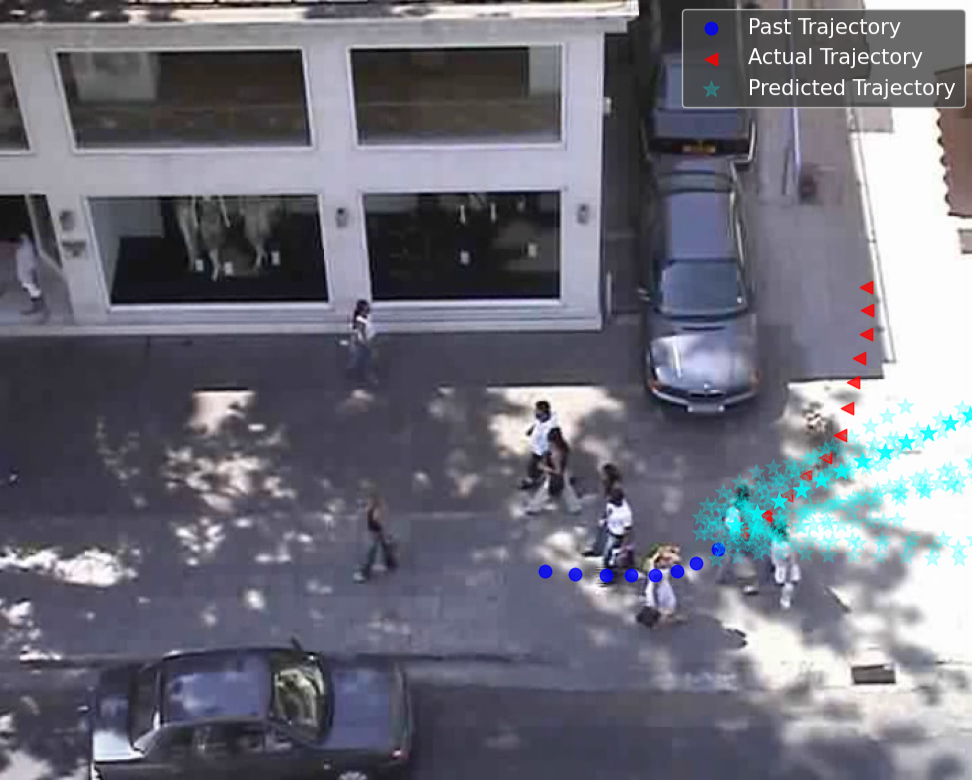}
        \caption{Zara01 (No Penalty)}
        \label{subfig:zara01_no_pen2}
    \end{subfigure}
    \hspace{0.1mm}
    \begin{subfigure}{0.475\textwidth}
        \centering
        \includegraphics[height=4.5cm,width=\columnwidth]{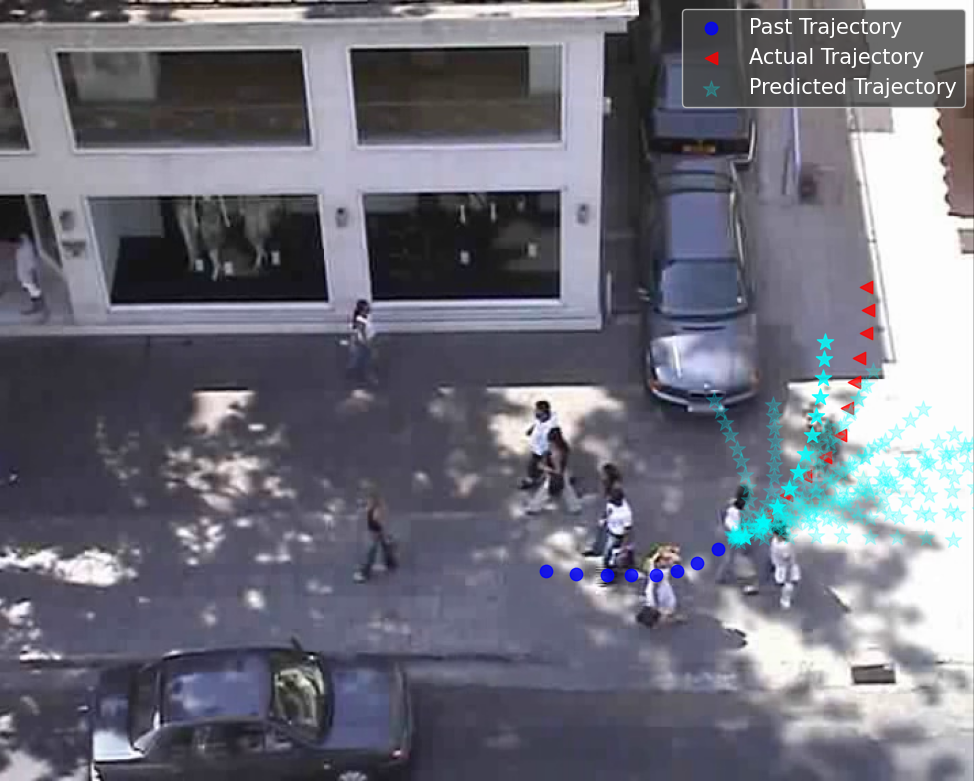}
        \caption{Zara01 (Penalty)}
        \label{subfig:zara01_pen2}
    \end{subfigure}
    \caption{Qualitative comparison of unpenalised vs. penalised trajectories on ETH-UCY dataset in stochastic setting.}
    \label{fig:penalized_vs_non}
\end{figure*}

\begin{table}[h]
\centering
\caption{\textbf{Ablation:} Comparing penalization strategies with SL1 loss on ETH-UCY (UNIV) dataset using ASTRA's SOTA configuration}
\vspace{-2mm}
\label{table:loss_p} 
{\small
\begin{tabular}{lcc}
\toprule
\textbf{Loss} & minADE$_{20}$/minFDE$_{20}$ \\ \midrule
\textbf{Unpenalised} & 0.37/0.52 \\
\textbf{Linear}  & 0.33/0.47 \\
\textbf{Quadratic} & 0.30/0.46 \\
\textbf{Parabolic} & \textbf{0.28/0.41} \\ \bottomrule
\end{tabular}
}
\end{table}

\begin{figure*}
    \centering
    % First row of images
    \begin{subfigure}{0.475\textwidth}
        \centering
        \includegraphics[height=4.5cm,width=\columnwidth]{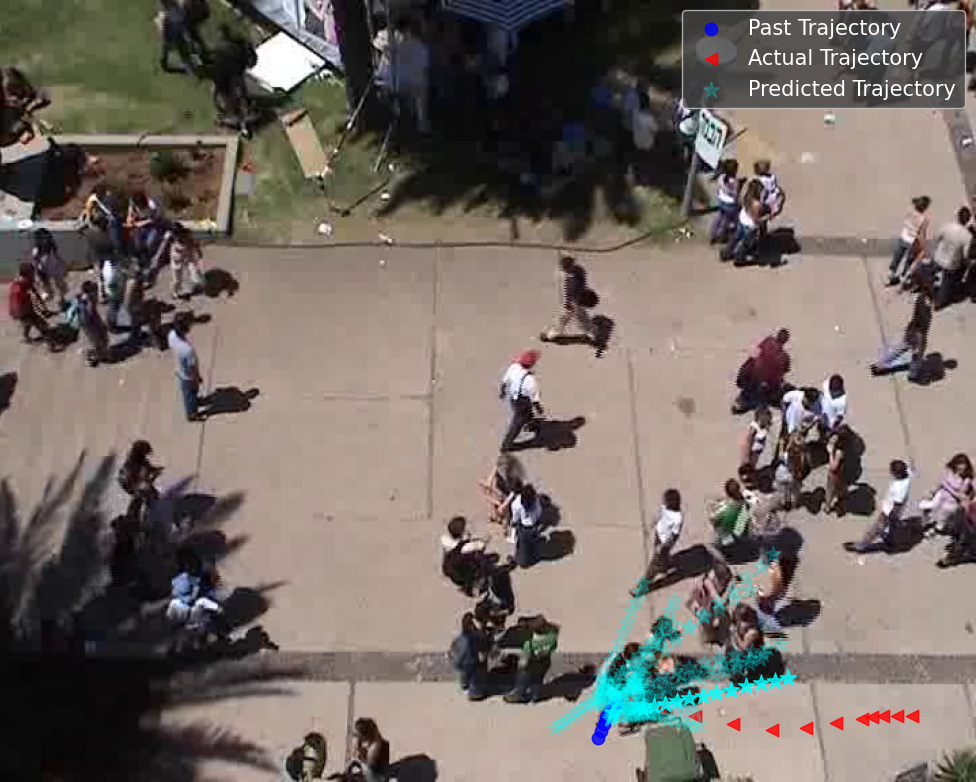}
        \caption{Univ (No Penalty)}
        \label{subfig:univ_no_pen2}
    \end{subfigure}
    \hspace{0.1mm}
    \begin{subfigure}{0.475\textwidth}
        \centering
        \includegraphics[height=4.5cm,width=\columnwidth]{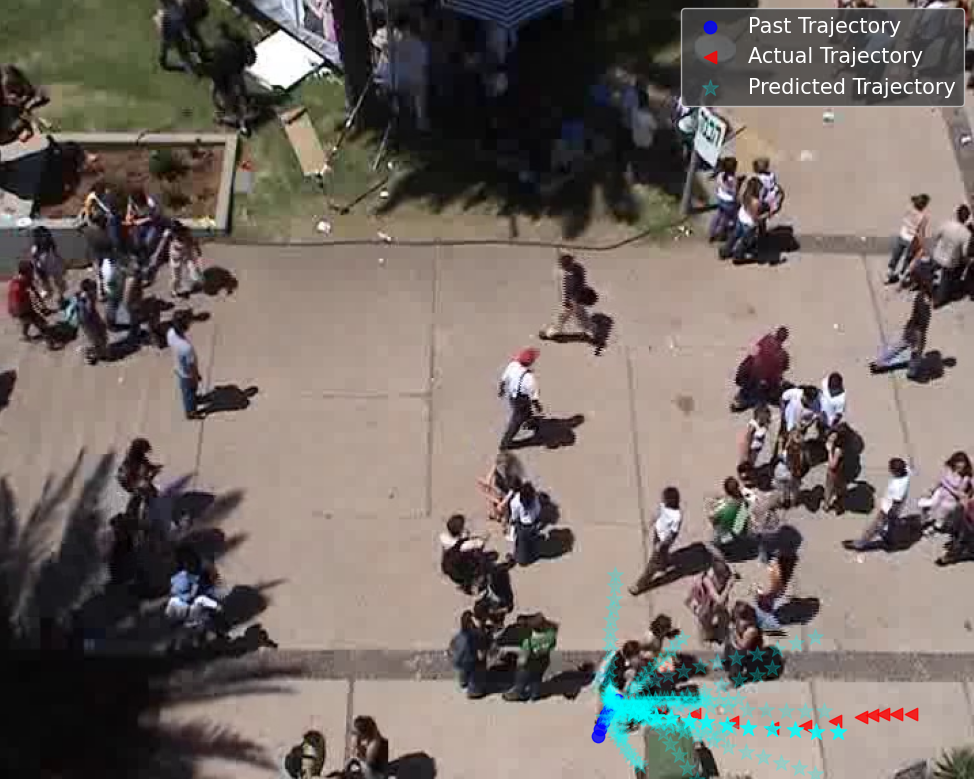}
        \caption{Univ (Penalty)}
        \label{subfig:univ_pen2}
    \end{subfigure}
    
    \vspace{5mm} % Space between rows

    \begin{subfigure}{0.475\textwidth}
        \centering
        \includegraphics[height=4.5cm,width=\columnwidth]{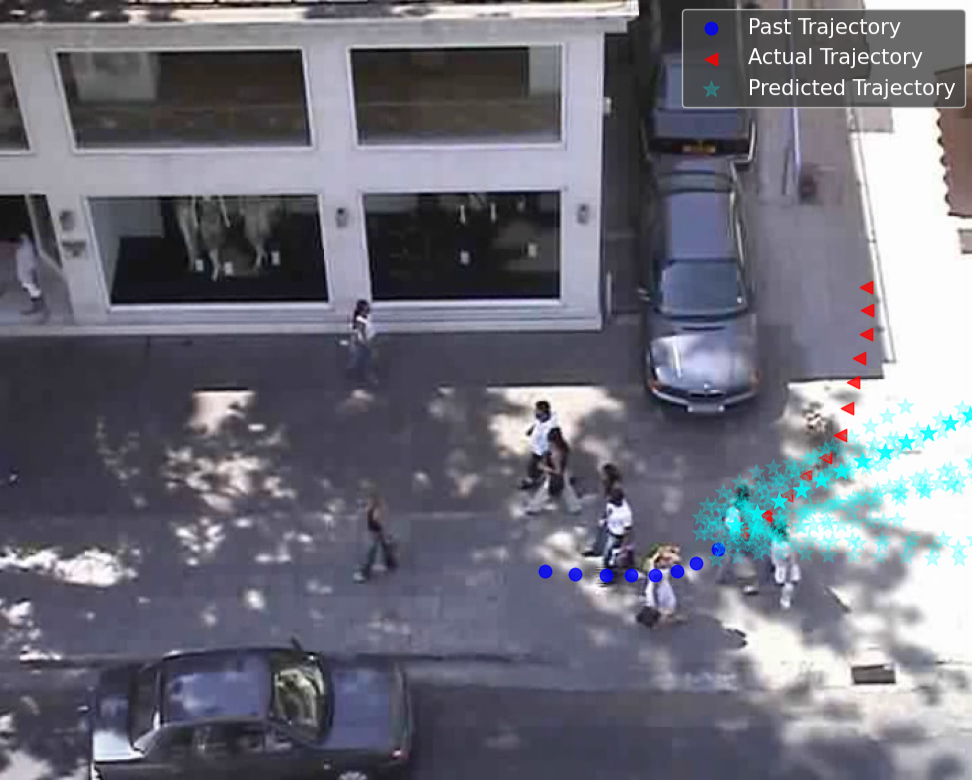}
        \caption{Zara01 (No Penalty)}
        \label{subfig:zara01_no_pen1}
    \end{subfigure}
    \hspace{0.1mm}
    \begin{subfigure}{0.475\textwidth}
        \centering
        \includegraphics[height=4.5cm,width=\columnwidth]{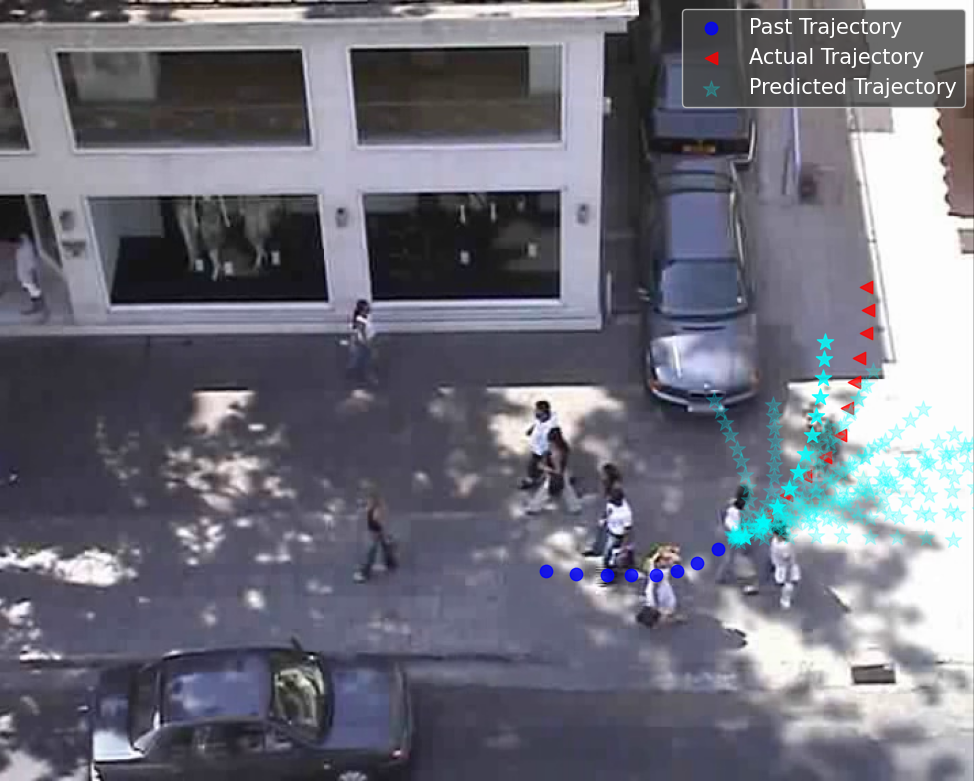}
        \caption{Zara01 (Penalty)}
        \label{subfig:zara01_pen1}
    \end{subfigure}
    \caption{Qualitative comparison of unpenalised vs. penalised trajectories on ETH-UCY dataset in stochastic setting.}
    \label{fig:penalized_vs_non}
\end{figure*}

% \begin{table}[h]
% \centering
% \caption{\textbf{Ablation:} Comparing penalization strategies with SL1 loss on ETH-UCY (UNIV) dataset using ASTRA's SOTA configuration}
% \vspace{-2mm}
% \label{table:loss_p} 
% {\small
% \begin{tabular}{lcc}
% \toprule
% \textbf{Loss} & minADE$_{20}$/minFDE$_{20}$ \\ \midrule
% \textbf{Unpenalised} & 0.37/0.52 \\
% \textbf{Linear}  & 0.33/0.47 \\
% \textbf{Quadratic} & 0.30/0.46 \\
% \textbf{Parabolic} & \textbf{0.28/0.41} \\ \bottomrule
% \end{tabular}
% }
% \end{table}

\subsection{Augmentation}
To enhance the robustness and generalization of our trajectory prediction model, we implement a data augmentation strategy. This strategy randomly applies rotation and translation transformations to the trajectory sequences \cite{zamboni2022pedestrian}. By applying random rotations and translations, our model is trained to be orientation-agnostic and adept at handling positional shifts in agents. These augmentations effectively increase the diversity of the training data, enabling the model to learn more generalized representations of agent movements. This, in turn, enhances the model's predictive accuracy and robustness, particularly in complex and unpredictable scenarios where pedestrian trajectories can vary significantly due to factors like group dynamics, obstacles, and varying crowd densities.

\section{Evaluation Metrics}
\textbf{ETH-UCY.} To evaluate our model on ETH-UCY, we used commonly employed evaluation metrics \cite{EqMotion,mohamed2020social,Chen2023vnagt,yu2020spatio}: ADE/FDE and minADE$_{K}$/minFDE$_{K}$. Average Displacement Error (ADE) computes the average Euclidean distance between the predicted trajectory and the true trajectory across all prediction time steps for each agent. 
$\text{minADE}_K$ refers to the minimum ADE out of K randomly generated trajectories and ground truth future trajectories.
We also used the Final Displacement Error (FDE), which focuses on the prediction accuracy at the final time step. It computes the Euclidean distance between the predicted and actual positions of each agent at the last prediction time step. $\text{minFDE}_K$ refers to the minimum FDE out of K randomly generated trajectories and ground truth future trajectories.
For multimodal trajectory prediction, $\text{minADE}_K$ and $\text{minFDE}_K$ metrics are used for evaluation.

\begin{equation}
\text{ADE} = \frac{1}{T_{pred}} \sum_{t=1}^{T_{pred}} \| Y^{a}_{t} - \hat{Y}^{a}_{t} \|_2.
\end{equation}

\begin{equation}
\text{minADE}_K = \min_{k} \left( \frac{1}{T_{pred}} \sum_{t=1}^{T_{pred}} \| Y^{a}_{t} - \hat{Y}^{a}_{t,k} \|_2 \right)
\end{equation}

\begin{equation}
\text{FDE} = \| Y^{a}_{T_{pred}} - \hat{Y}^{a}_{T_{pred}} \|_2
\end{equation}

\begin{equation}
\text{minFDE}_K = \min_{k} \left( \| Y^{a}_{T_{pred}} - \hat{Y}^{a}_{T_{pred},k} \|_2 \right)
\end{equation}

\textbf{PIE.} For the PIE Dataset, the ADE and FDE metrics are calculated based on the centroid of the bounding box \cite{piedataset,rasouli2023pedformer}, denoted as Centre average displacement error for the bounding box (CADE) and Centre final displacement error for the bounding box (CFDE). 
In addition, we reported the average and final Root Mean Square Error (RMSE) of bounding box coordinates, denoted as ARB and FRB, respectively \cite{Rasouli2021}. 
 
\section{Grad-CAM visualizations}
Grad-CAM images were obtained by generating heatmaps overlaid onto the original image to aid in validating the relevance of highlighted regions. To obtain the Grad-CAM visualization, a single-channel output segmentation map was obtained from the pre-trained U-Net network, representing the probability of each pixel location being a keypoint \cite{ribera2019locating}. Probabilities were aggregated across all pixels, by comparing them with true keypoints and gradients of activation for the initial layer were extracted, similar to the approach taken by \cite{Vinogradova_2020}. Utilizing these gradients, a weighted average of the activation maps of the initial layer was computed to reconstruct the heatmap, similar to the method described in \cite{selvaraju2017grad}, for the Grad-CAM visualization. Overlaying this heatmap onto the original image highlights the regions that contribute significantly to the keypoint predictions made by the model.

\begin{figure*}
    \centering
    % First row of images
    \begin{subfigure}{0.475\textwidth}
        \centering
        \includegraphics[height=4.5cm,width=\columnwidth]{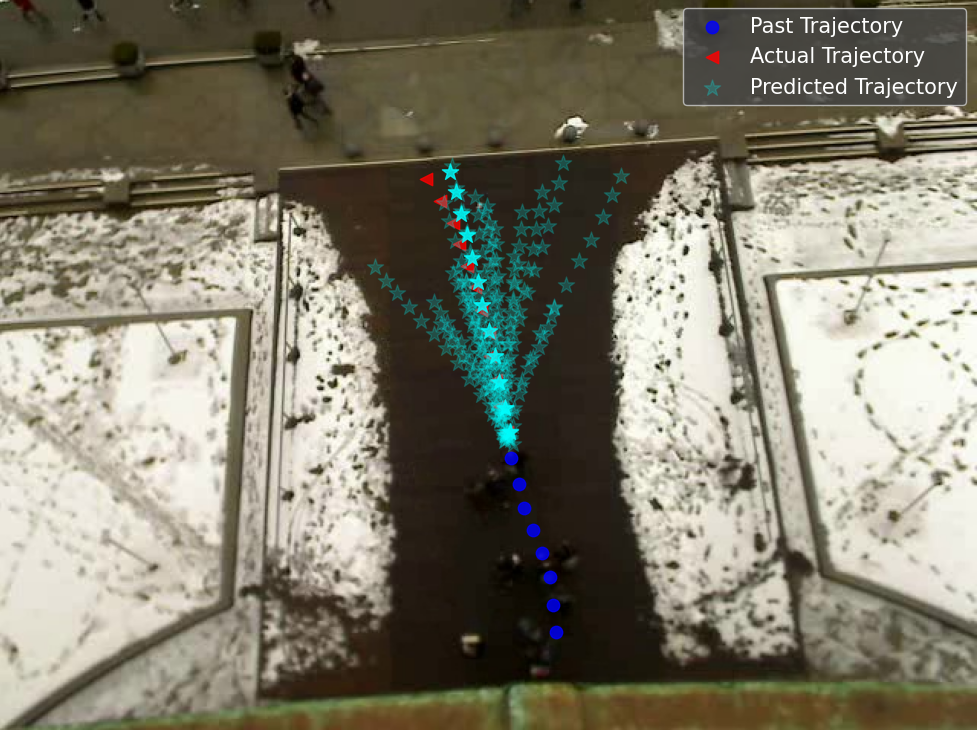}
        \caption{ETH}
        \label{subfig:eth_st}
    \end{subfigure}
    \hspace{0.1mm}
    \begin{subfigure}{0.475\textwidth}
        \centering
        \includegraphics[height=4.5cm,width=\columnwidth]{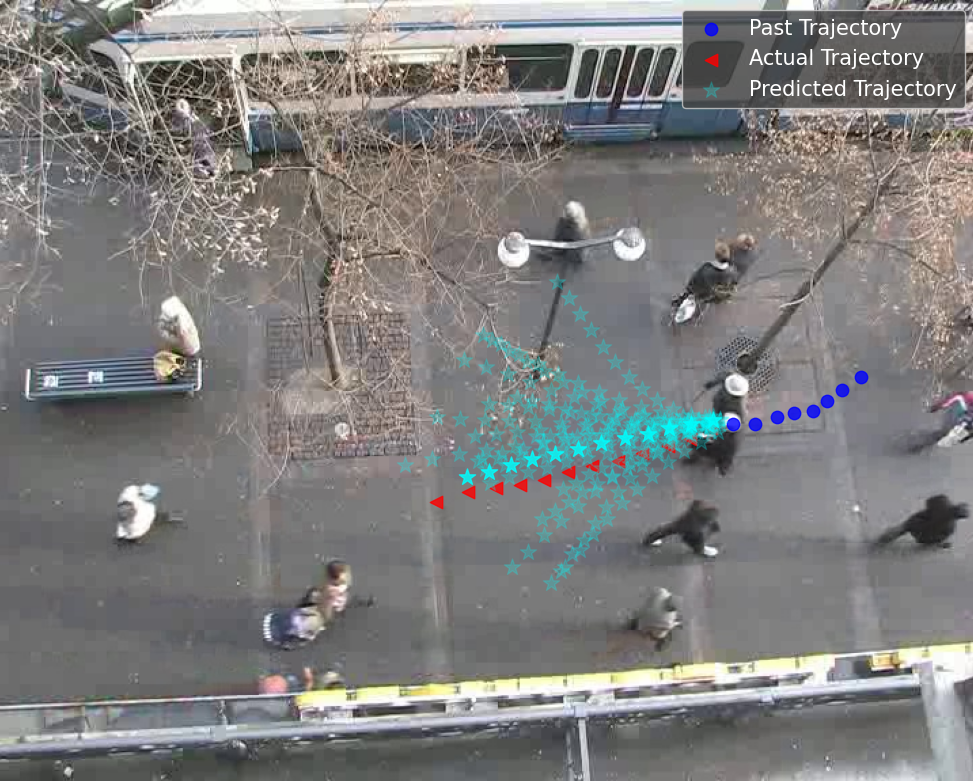}
        \caption{Hotel}
        \label{subfig:hotel_st}
    \end{subfigure}
    
    \vspace{5mm} % Space between rows

    % Second row of images
    \begin{subfigure}{0.475\textwidth}
        \centering
        \includegraphics[height=4.5cm,width=\columnwidth]{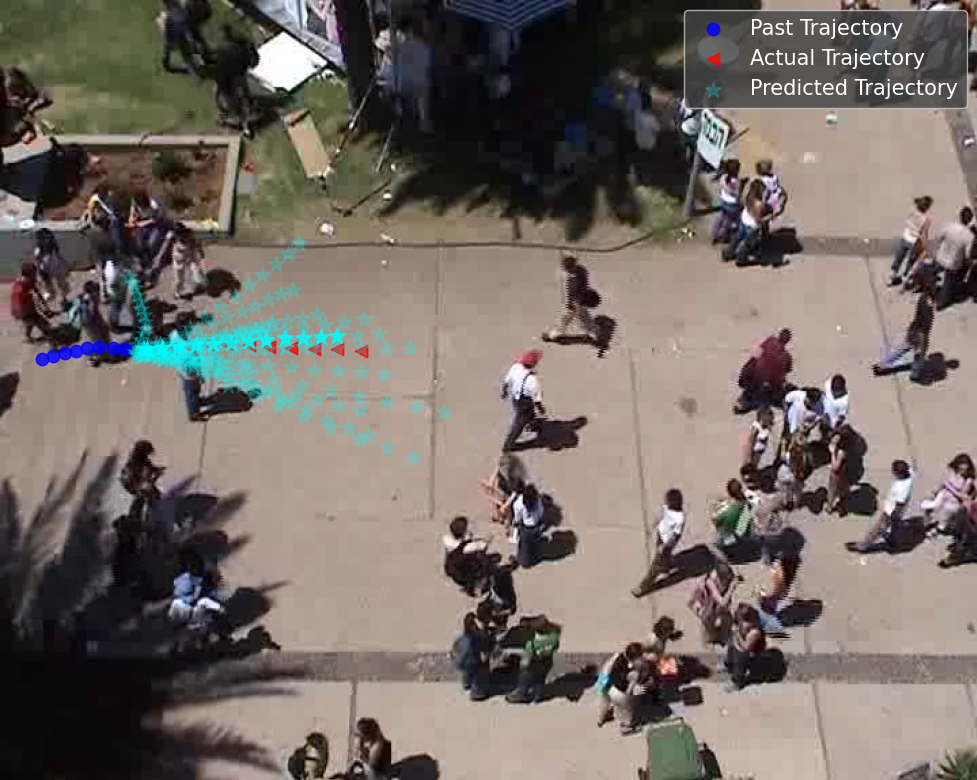}
        \caption{Univ}
        \label{subfig:univ_st}
    \end{subfigure}
    \hspace{0.1mm}
    \begin{subfigure}{0.475\textwidth}
        \centering
        \includegraphics[height=4.5cm,width=\columnwidth]{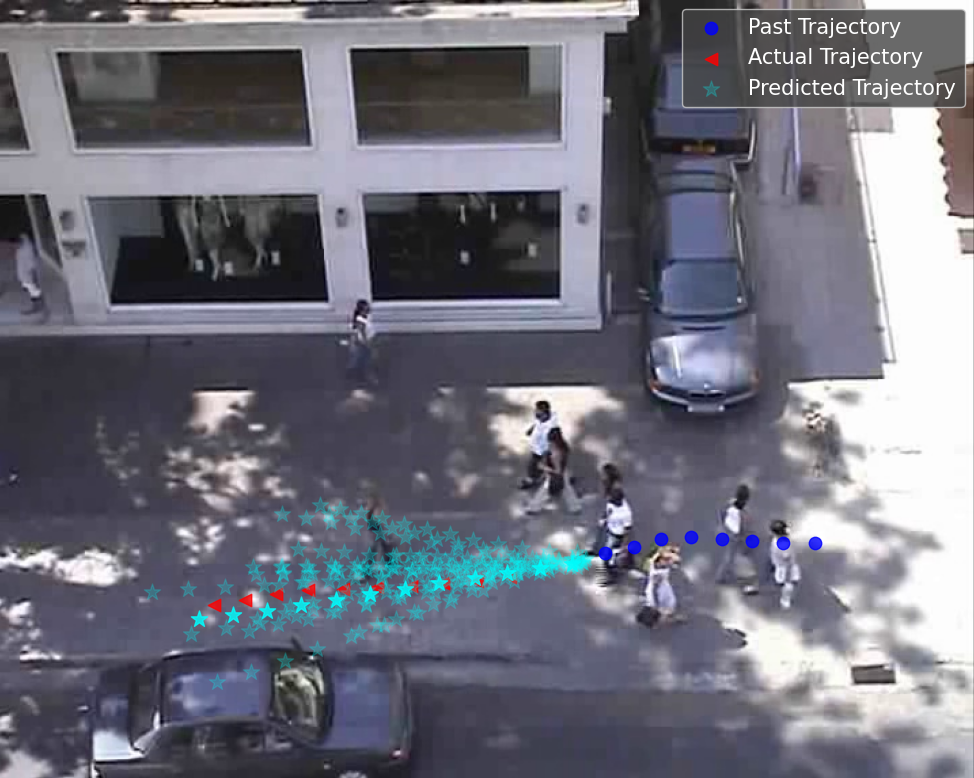}
        \caption{Zara1}
        \label{subfig:zara01_st}
    \end{subfigure}
    
    \vspace{5mm} % Space between rows

    % Third row of images
    \begin{subfigure}{0.475\textwidth}
        \centering
        \includegraphics[height=4.5cm,width=\columnwidth]{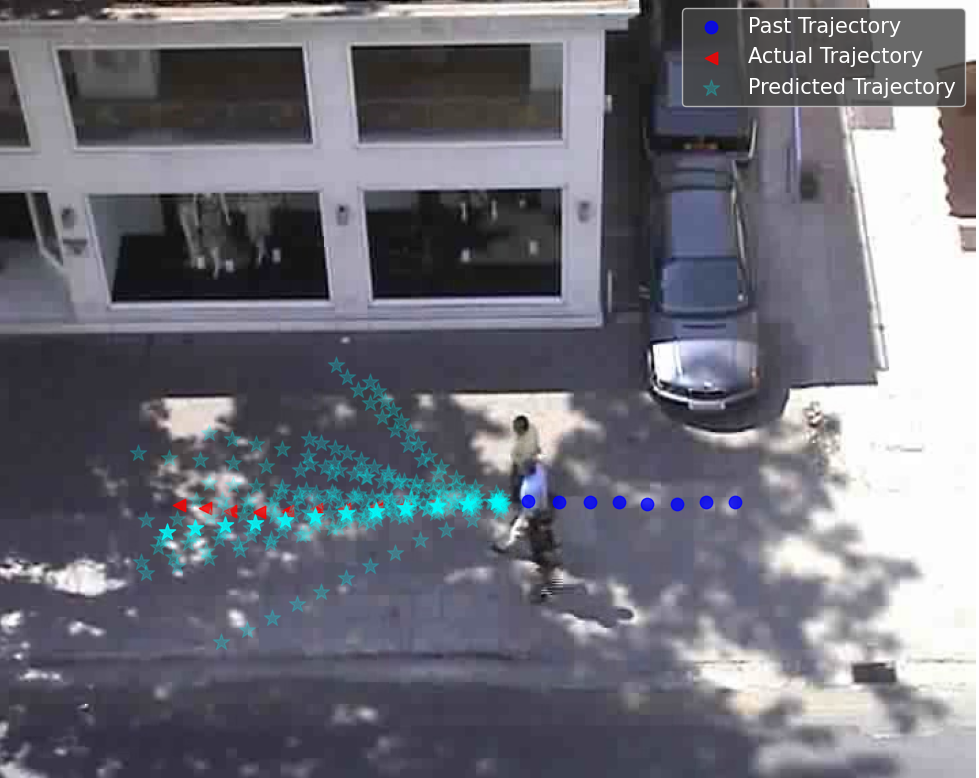}
        \caption{Zara2}
        \label{subfig:zara02_st}
    \end{subfigure}
    % \hspace{0.1mm}
    % \begin{subfigure}{0.475\textwidth}
    %     \centering
    %     \includegraphics[height=6cm,width=\columnwidth]{Figures/supp_imgs/pie.png}
    %     \caption{Caption 6}
    %     \label{subfig:6}
    % \end{subfigure}

    \caption{Multi-modal trajectory visualizations on ETH-UCY dataset (BEV)}
    \label{fig:stochastic_result}
\end{figure*}

\begin{figure*}
    \centering
    % First row of images
    \begin{subfigure}{0.475\textwidth}
        \centering
        \includegraphics[height=4.5cm,width=\columnwidth]{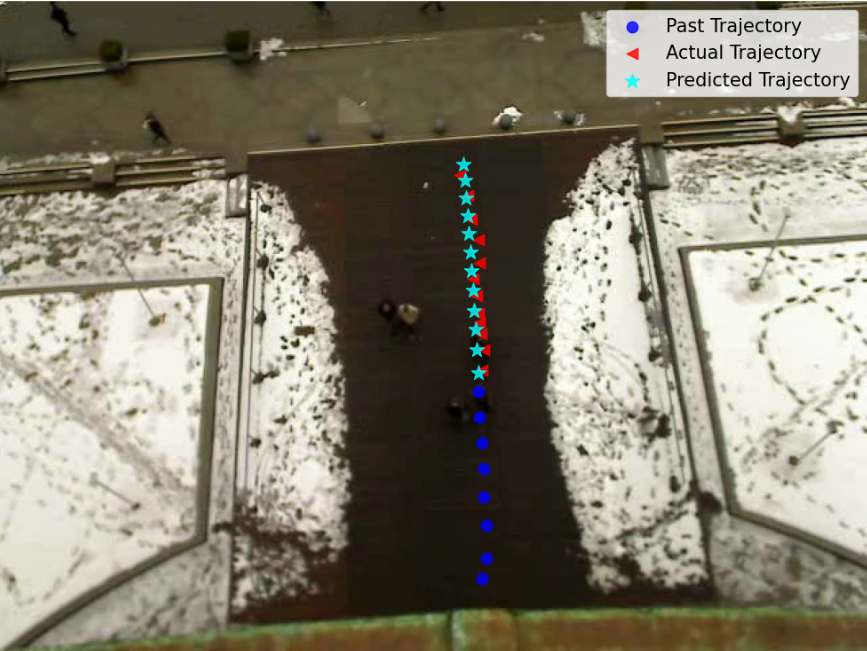}
        \caption{ETH}
        \label{subfig:1}
    \end{subfigure}
    \hspace{0.1mm}
    \begin{subfigure}{0.475\textwidth}
        \centering
        \includegraphics[height=4.5cm,width=\columnwidth]{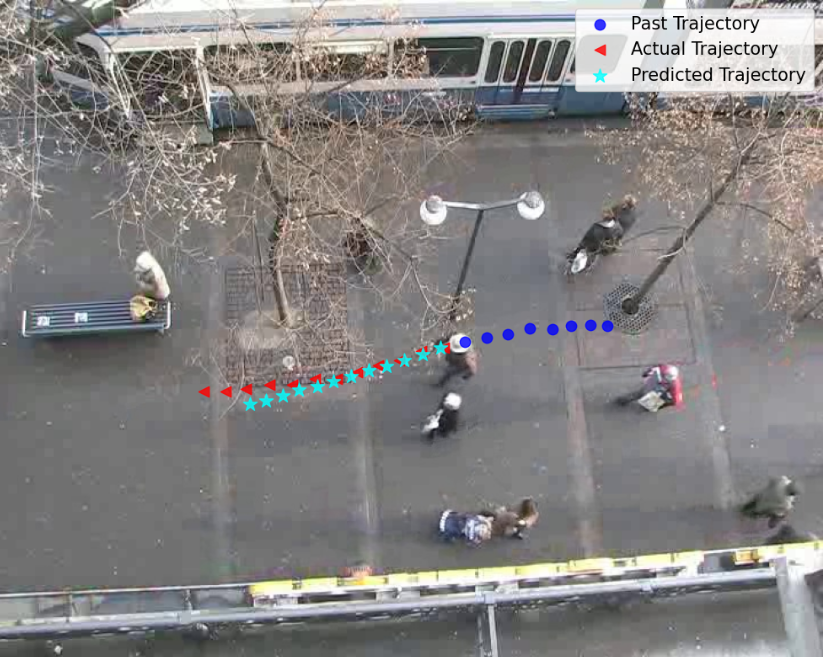}
        \caption{Hotel}
        \label{subfig:2}
    \end{subfigure}
    
    \vspace{5mm} % Space between rows

    % Second row of images
    \begin{subfigure}{0.475\textwidth}
        \centering
        \includegraphics[height=4.5cm,width=\columnwidth]{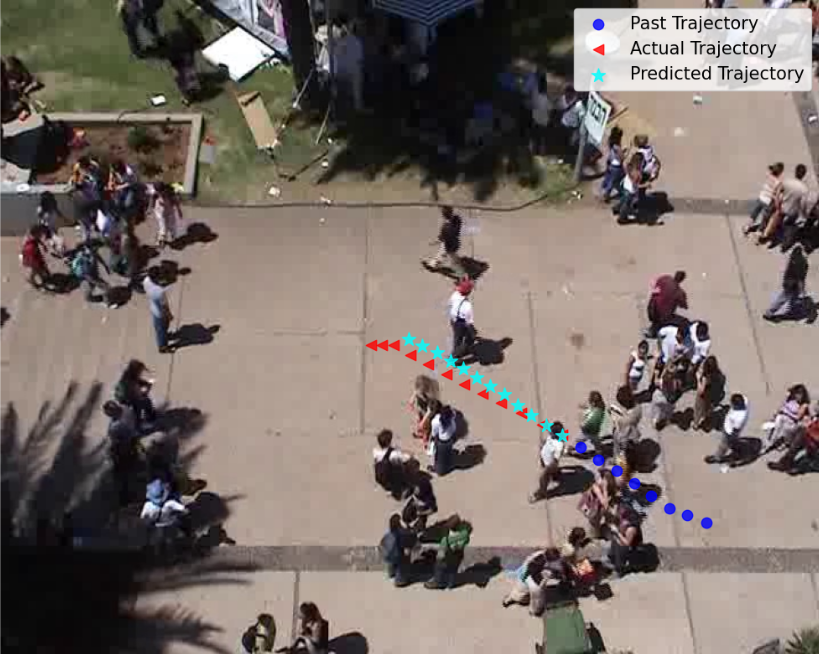}
        \caption{Univ}
        \label{subfig:3}
    \end{subfigure}
    \hspace{0.1mm}
    \begin{subfigure}{0.475\textwidth}
        \centering
        \includegraphics[height=4.5cm,width=\columnwidth]{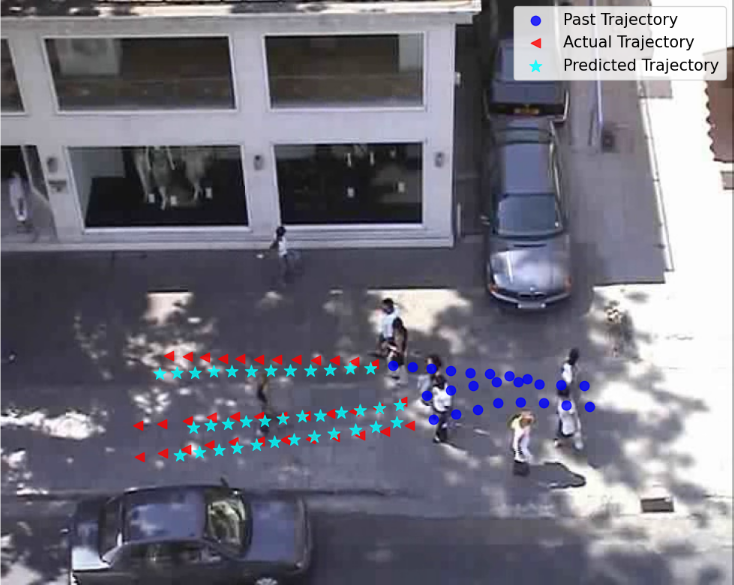}
        \caption{Zara1}
        \label{subfig:4}
    \end{subfigure}
    
    \vspace{5mm} % Space between rows

    % Third row of images
    \begin{subfigure}{0.475\textwidth}
        \centering
        \includegraphics[height=4.5cm,width=\columnwidth]{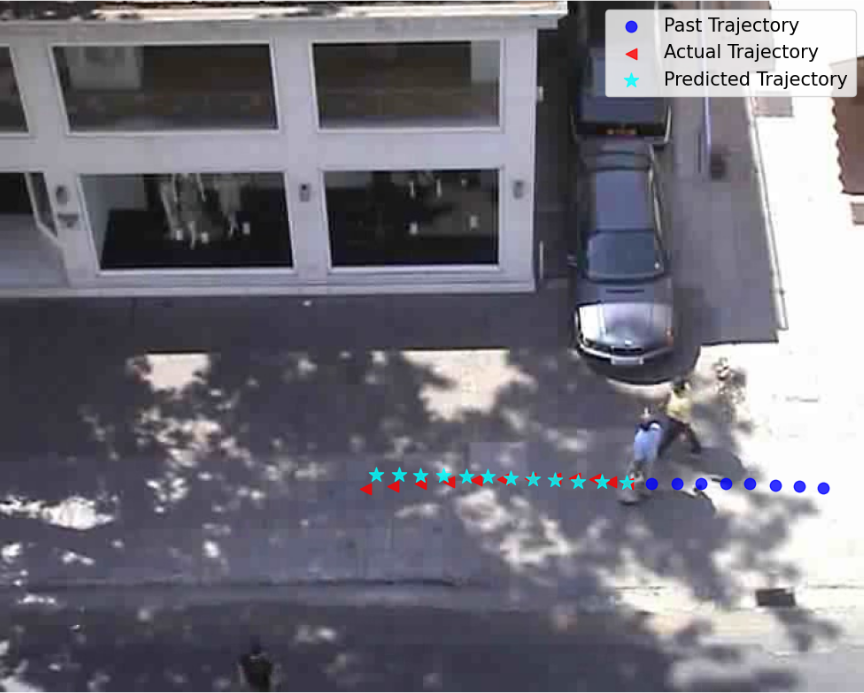}
        \caption{Zara2}
        \label{subfig:5}
    \end{subfigure}
    % \hspace{0.1mm}
    % \begin{subfigure}{0.475\textwidth}
    %     \centering
    %     \includegraphics[height=6cm,width=\columnwidth]{Figures/supp_imgs/pie.png}
    %     \caption{Caption 6}
    %     \label{subfig:6}
    % \end{subfigure}

    \caption{Deterministic trajectory visualizations on ETH-UCY dataset (BEV)}
    \label{fig:unimodal_result}
\end{figure*}

\begin{figure*}[ht]
    \centering
    % First image
    \begin{subfigure}{0.475\textwidth}
        \centering
        \includegraphics[height=5cm, width=\linewidth]{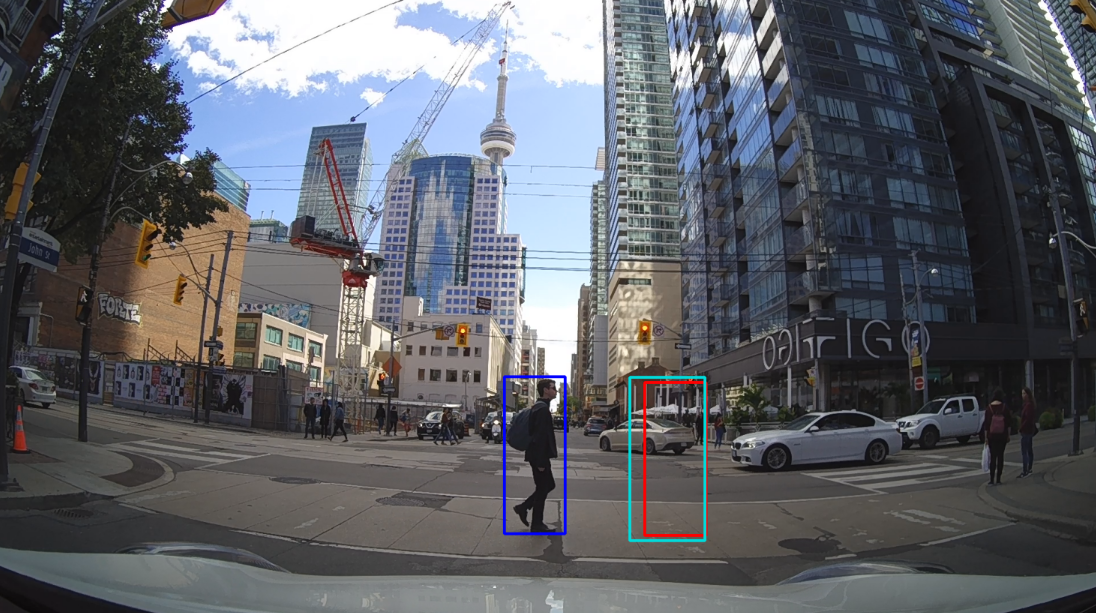}
        \caption{}
        \label{subfig:first}
    \end{subfigure}
    \hspace{0.1mm} % Adjust spacing between images as needed
    % Second image
    \begin{subfigure}{0.475\textwidth}
        \centering
        \includegraphics[height=5cm, width=\linewidth]{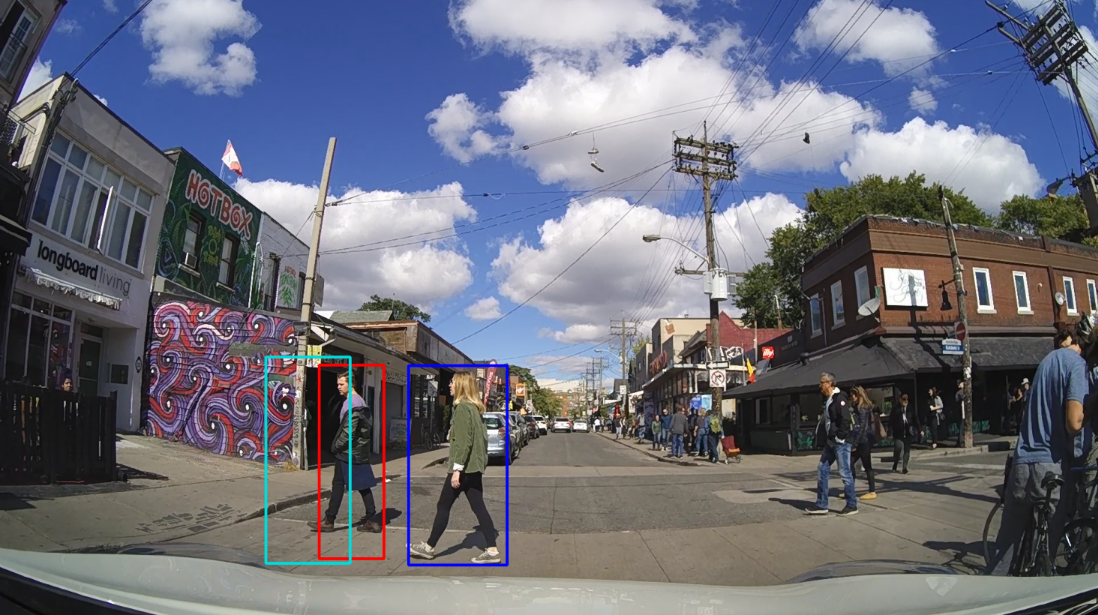}
        \caption{}
        \label{subfig:second}
    \end{subfigure}
    
    \caption{Trajectory Visualizations on PIE Dataset (EVV) 
where the red and cyan bounding box indicates the ground-truth and predicted final position respectively and the blue bounding box indicates the start position.}
\end{figure*}

To ease reading the paper, Table \ref{tab:abbrv} and \ref{tab:symb} list the abbreviations and the mathematical symbols mentioned in the paper, respectively. 

\begin{table*}[!ht]
    \centering
    \caption{Table of Abbreviations Used}
    
    \label{tab:abbrv}
    \begin{tabular}{ll}
        \noalign{\hrule height 1pt}
        \textbf{Abbreviation/Term} & \textbf{Description} \\
        \hline
        ASTRA & Agent-Scene aware model for pedestrian trajectory forecasting\\
        BEV & Bird's Eye View \\
        EVV & Ego-Vehicle View \\
        AV & Autonomous Vehicle \\
        MLP & Multi-Layer Perceptron \\
        % GCN & Graph Convolutional Network \\
        CVAE & Conditional Variational Auto-Encoder\\
        GNN & Graph Neural Network \\
        RWPE & Random Walk Positional Encoding \\
        MSE & Mean Square Error (Loss Function) \\
        SL1 & Smooth L1 Loss (Loss Function) \\
        ADE                 & Average Displacement Error \\ 
FDE                 & Final Displacement Error \\ 
CADE & Centre average displacement error for the bounding box \\
CFDE & Centre final displacement error for the bounding box \\   
ARB & Average Root Mean Square Error for the bounding box\\
FRB & Final Root Mean Square Error for the bounding box \\
        \noalign{\hrule height 1pt}
    \end{tabular}
\end{table*}

\begin{table*}
    \centering
    \caption{Table of Mathematical Symbols Used}
    
    \label{tab:symb}
    \begin{tabular}{ll}
        \noalign{\hrule height 1pt}
        \textbf{Symbols} & \textbf{Description} \\
        \hline   
$N$                 & Total number of predictions in MSE calculation \\ 
        $\boldsymbol{X}$ & Observed trajectories of agents \\
        $\boldsymbol{Y}$ & Groundtruth future trajectories of agents \\
        $\boldsymbol{\hat{Y}}$ & Predicted trajectories of agents \\
        % $\gamma$ & \\
        $T_{\text{obs}}$ & Number of past time instants for observation \\
        $T_{\text{pred}}$ & Number of future time instants for prediction \\
        $\boldsymbol{I_{t=1:T_{\text{Obs}}}}$ & Sequence of past input frame images \\
        $X^{a}_{t}$         & Observed coordinates for agent $a$ at time $t$ \\ 
         $\hat{Y}^{a}_t$ & Predicted coordinates for agent $a$ at time $t$ \\
$A$                 & Number of target agents \\ 
        $e_{ij}$            & Edge weight in graph $G$ between nodes $i$ and $j$ \\ 
       
$d(v_i, v_j)$       & Distance between agents $v_i$ and $v_j$ \\ 
$w(t)$              & Weight function in weighted-penalty strategy \\ 
$w_{\text{start}}$  & Start weight in weighted-penalty strategy \\ 
$w_{\text{end}}$    & End weight in weighted-penalty strategy \\

        $\boldsymbol{\Psi_{\text{Scene}}}$ & Latent representation of scene(past frame) obtained from U-Net encoder \\ 
        $\boldsymbol{\Phi_{\text{Scene}}}$ & Scene-aware embeddings \\
        $\boldsymbol{T_{\text{Scene-aware}}}$ & Scene-aware Transformer encoder \\
        $\boldsymbol{\Upsilon_{\text{Encoder}}}$ & U-Net Encoder \\
        $\boldsymbol{\Gamma_{\text{Scene}}}$ & Multi-layer Perceptron layer for Scene embeddings \\

        $\boldsymbol{\Phi_{\text{Temporal}}}$ & Temporal encoding \\

        $\boldsymbol{\Gamma_{\text{Spatial}}}$ & Multi-layer Perceptron layer for Spatial embeddings \\
        $\boldsymbol{\Phi_{\text{Spatial}}}$ & Spatial embeddings \\
        $\boldsymbol{\Gamma_{\text{Social}}}$ & Multi-layer Perceptron layer for Social embeddings \\
        $\boldsymbol{\Phi_{\text{Social}}}$ & Social Embeddings \\

        $\boldsymbol{T_{\text{Agent-aware}}}$ & Agent-aware Transformer encoder \\
        $\boldsymbol{\Phi_{\text{Agents}}}$ & Agent-aware embeddings \\
        $L_{\text{weighted}}(\hat{Y}, Y)$ & Weighted-penalty Loss Function \\
        \noalign{\hrule height 1pt}
    \end{tabular}
\end{table*}

\end{document}

%% file: math_commands.tex
\usepackage{amsmath,amsfonts,bm}

\def\eqref#1{equation~\ref{#1}}
\def\1{\bm{1}}

\DeclareMathAlphabet{\mathsfit}{\encodingdefault}{\sfdefault}{m}{sl}
\SetMathAlphabet{\mathsfit}{bold}{\encodingdefault}{\sfdefault}{bx}{n}